%% file: main.tex
\documentclass[10pt, conference]{IEEEtran}

\usepackage{amsmath,amssymb,amsfonts}
\usepackage{graphicx}
\usepackage{pifont}

\usepackage{amsthm}
\usepackage{xcolor}
\usepackage[figuresright]{rotating}

\usepackage{graphicx}
\graphicspath{{/}{fig/}}

\usepackage{array}
\usepackage{textcomp}
\usepackage{xcolor}
\usepackage{multirow}
\usepackage{booktabs}

\usepackage{mathtools}
\usepackage{breqn}
\usepackage{float}

\usepackage{pgfplots}
\pgfplotsset{compat=newest}
\usepgfplotslibrary{groupplots}

\usepackage{caption}
\usepackage{subcaption}

\usepackage{multirow,tabularx}
\usepackage{hyperref}
\usepackage{flushend}
\usepackage{algorithmic}
\usepackage[vlined, ruled, shortend]{algorithm2e}

\newlength\figureheight
\newlength\figurewidth
\SetAlCapNameFnt{\footnotesize}
\SetAlCapFnt{\footnotesize}

\title{
    Benchmarking ML Approaches to UWB-Based Range-Only Posture Recognition for \\ Human Robot-Interaction
}

\author{
    \IEEEauthorblockN{
        \vspace{1em}
        Salma Salimi\IEEEauthorrefmark{1},
        Sahar Salimpour\IEEEauthorrefmark{1},
        Jorge Pe\~na Queralta\IEEEauthorrefmark{2},
        Wallace Moreira Bessa\IEEEauthorrefmark{1},
        Tomi Westerlund\IEEEauthorrefmark{1}
    }
    \IEEEauthorblockA{
        \normalsize
        \IEEEauthorrefmark{1}\href{https://tiers.utu.fi}{Faculty of Technology, University of Turku, Finland}\\
        \IEEEauthorrefmark{2}\href{https://tiers.utu.fi}{Swiss Federal School of Technology, ETH Zürich,Switzerland}\\
        [+6pt]
    }
}

\begin{document}

\maketitle
\thispagestyle{empty}
\pagestyle{empty}

\input{sec/00_Abstract.tex}
\IEEEpeerreviewmaketitle

\input{sec/01_Intro}
\input{sec/02_Methods}

\input{sec/03_ModelPerformance}

\input{sec/04_Experiments}
\input{sec/05_Discussion}
\input{sec/06_Conclusion}




\bibliographystyle{unsrt}
\bibliography{bibliography}

\end{document}

%% file: sec/00_Abstract.tex

\begin{abstract}%
    \label{sec:abstract}%
Human pose estimation involves detecting and tracking the positions of various body parts using input data from sources such as images, videos, or motion and inertial sensors. This paper presents a novel approach to human pose estimation using machine learning algorithms to predict human posture and translate them into robot motion commands using ultra-wideband (UWB) nodes, as an alternative to motion sensors. The study utilizes five UWB sensors implemented on the human body to enable the classification of still poses and more robust posture recognition. This approach ensures effective posture recognition across a variety of subjects. These range measurements serve as input features for posture prediction models, which are implemented and compared for accuracy. For this purpose, machine learning algorithms including K-Nearest Neighbors (KNN), Support Vector Machine (SVM), and deep Multi-Layer Perceptron (MLP) neural network are employed and compared in predicting corresponding postures. We demonstrate the proposed approach for real-time control of different mobile/aerial robots with inference implemented in a ROS~2 node. Experimental results demonstrate the efficacy of the approach, showcasing successful prediction of human posture and corresponding robot movements with high accuracy.

\end{abstract}


%% file: sec/01_Intro.tex

 \section{Introduction}\label{sec:introduction}

In the current rapidly evolving technological landscape, robotics has assumed a pivotal role across diverse domains, from industrial automation and healthcare to entertainment and everyday tasks. As part of this integration, Human-Robot Interaction (HRI) facilitates the deployment of robots in human environments, spanning collaborative work settings such as manufacturing~\cite{kumar2020survey}, service robots~\cite{mohebbi2020human}, search and rescue robots, and drones~\cite{lyu2023unmanned}. Robots with advanced sensors and AI, controlled by humans, can handle precise and strenuous tasks, while humans focus on decision-making.

Within HRI, the ability of robots to accurately perceive and interpret a wide range of human communication methods is essential~\cite{su2023recent}. Establishing a seamless communication system between humans and robots has emerged as a key research focus. When verbal commands or even neural signals are not the preferred mode of communication, human postures serve as an important channel for conveying intentions and commands~\cite{neto2019gesture}. Through body postures, individuals can convey discrete but precise signals to the robot to execute predefined tasks. Vision-based and wearable-based approaches dominate this field when it comes to creating posture-based interaction interfaces~\cite{villani2023general}.



Traditional human pose estimation methods predominantly relied on computer vision techniques, utilizing cameras such as RGB or infrared~\cite{kim2023study}. These systems face challenges like complicated backgrounds, occlusions, and low-resolution images. In recent years, there has been a significant trend towards 3D pose estimation. Accurate 3D pose annotations are often costly because they require specialized systems like motion capture with markers on the subject. Consequently, researchers have concentrated on creating techniques to derive 3D human pose data from 2D images. However, these 3D annotations are typically obtained in controlled environments, limiting their applicability in real-world scenarios. Some techniques have attempted to address this by using lifting operations to estimate 3D poses from 2D images captured by cameras~\cite{gosztolai2020liftpose3d}. Despite this, vision-based methods struggle with occlusions and depth ambiguity. For instance, transformer-based approaches have been used to mitigate partial occlusions by utilizing attention modules to capture global context and long-range relationships between the predicted joints~\cite{dosovitskiy2020image}. Nonetheless, these methods still face challenges in scenarios with complete occlusion. In addition to vision-based systems, IMU sensors have been used to measure motion. However, IMU sensors can become inaccurate with movement, making it difficult to classify static postures. This inaccuracy further complicates the task of human posture estimation.

Recently, Ultra-Wide-Band (UWB) technology has emerged as a promising alternative due to its low complexity, high accuracy, and ability to penetrate obstacles~\cite{martinelli2023unposed}. UWB technology offers a robust solution to the limitations of vision-based systems and IMUs. One implementation of UWB involves an impulse radio ultra-wideband (IR-UWB) radar system with an 8-by-8 multiple-input multiple-output (MIMO) antenna array, suitable for through-wall detection. The 3D-TransPOSE algorithm~\cite{kim2023study}, based on transformer architecture, improves accuracy by focusing on relevant radar signal segments. However, it has limited penetration depth and complex hardware requirements. In contrast, methods using camera-equipped flying robots~\cite{nageli2018flycon} offer real-time motion tracking but face challenges like high computational demands and safety concerns.

Over the past decade, UWB has evolved into a dependable, cost-effective, and commercially viable RF solution for data transmission, Time of Flight (TOF) or Time Difference of Arrival (TDOA) ranging, and localization~\cite{fishberg2022multi}. Consequently, many roboticists have started integrating UWB into their projects due to its precision of approximately 10 centimeters, range of up to 100 meters, resilience to multipath interference, independence from line of sight (LOS), low power consumption, and communication speeds of 100 Mbit/s~\cite{alarifi2016ultra}. Despite these advantages, UWB measurements can still be affected by ranging errors and noise, which remains an active area of research in the robotics community~\cite{smaoui2020study,hamer2018self,ledergerber2018calibrating}.

In this paper, we propose utilizing UWB technology combined with transformer-based algorithms for human posture estimation. The micro-Doppler radar effect, for instance, has been employed to classify human movement without estimating body parts in~\cite{du2019three}, and a bistatic radar configuration using UWB has been developed to locate a moving person by analyzing the channel impulse response in~\cite{doglioni2022cost}. Additionally, RF-Pose3D has shown the potential of using RF signals to infer 3D human poses, even with multiple individuals and obstacles such as walls and occlusions~\cite{zhao2018rf}. By leveraging these advances, our approach aims to overcome the limitations of traditional methods and provide more accurate and reliable human posture estimation.

We employ UWB sensors in our study for human posture estimation, a novel application to the best of our knowledge. By utilizing node distances placed on the human body, we aim to generate precise and reliable pose predictions. To validate our approach, we propose incorporating machine learning algorithms such as KNN, SVM, and MLP. Our model will be trained on UWB data to accurately predict human posture, allowing us to issue commands to robots based on these predictions.

Our approach offers several key contributions. Firstly, it provides enhanced accuracy, offering stable and accurate pose estimation regardless of motion. Secondly, it overcomes the limitations of traditional methods; unlike camera-based systems, our approach does not struggle with occlusions and depth ambiguity. Thirdly, UWB sensors are less demanding in terms of computational resources and power, making our approach more efficient. Finally, we provide a comprehensive dataset in our GitHub to support further research in the field, enhancing accessibility and reproducibility.

The remainder of this paper is organized as follows: Section II details the methodology and background on employing different machine learning techniques with UWB technology for human pose estimation. Section III evaluates the classification performance of the various models introduced in Section II. Section IV presents the experimental results, and finally, Section V offers the conclusion.

%% file: sec/02_Methods.tex
\section{Methodology} \label{sec:methods}

\begin{figure*}[tb]
    \centering
    \includegraphics[width=1.0\textwidth]{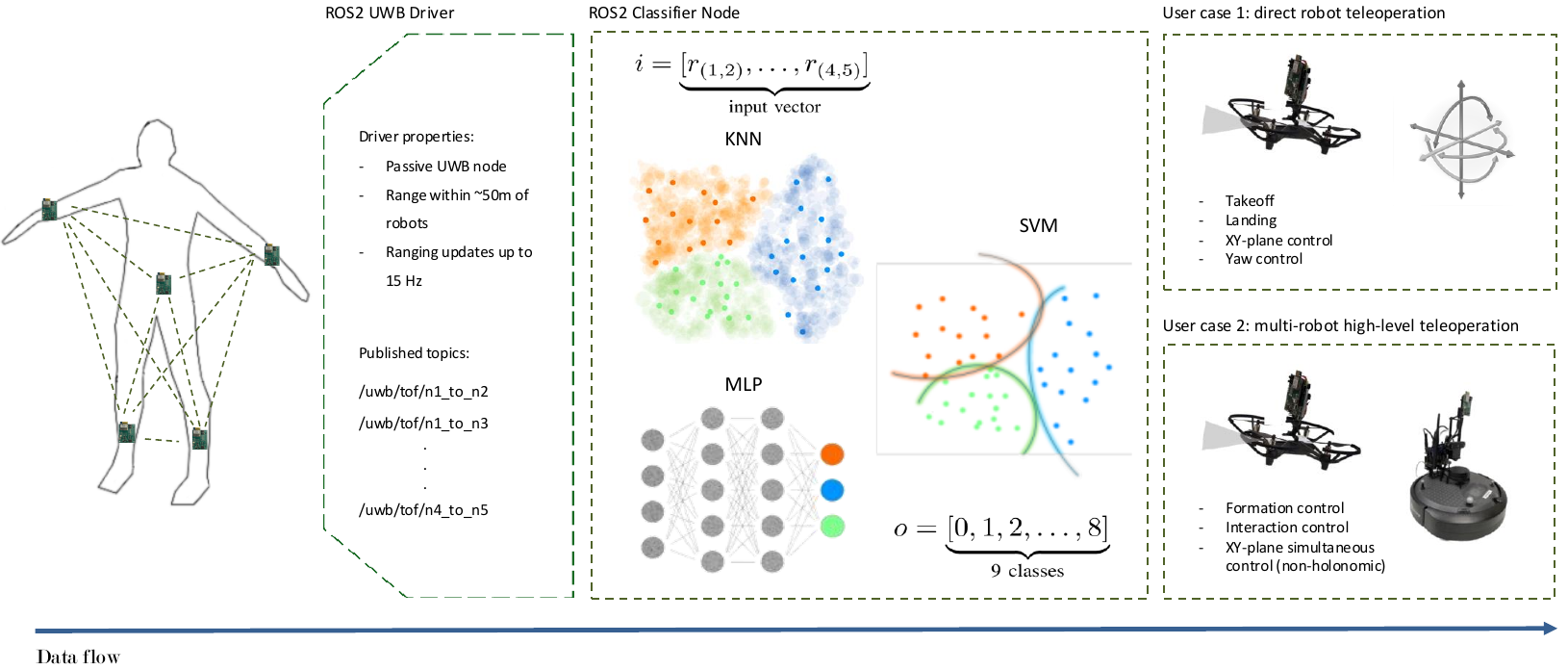}
    \caption{Illustration of the blocks, data flow and use-cases in the presented work. Up to 5 UWB nodes are used to estimate a set of predefined human motions for controlling a single aerial robot or a multi-robot system. We compare the performance of different machine learning algorithms for such task.}
    \label{fig:diagram}
\end{figure*}

We performed human pose estimation using UWB sensor data by applying and evaluating various machine learning algorithms as illustrated in Fig.~\ref{fig:diagram}. 
The following sections delve into each step with comprehensive elaboration.

\subsection{UWB Ranging}
The high-accuracy short-range distance between two UWB nodes can be estimated using the ToF of a wireless signal exchanged between them, multiplied by the speed of light. ToF can be measured using one-way or two-way ranging, depending on device synchronization. One-way ranging requires precise clock synchronization as the signal travels in only one direction, whereas two-way ranging does not need synchronization, making it more commonly used due to its simplicity~\cite{queralta2020uwb}. In this paper, we use Single-Sided Two-Way Ranging (SS-TWR), where the distance is calculated based on the time of flight of a UWB signal
query and its reply. Errors in range estimates between UWB nodes, which depend on different environmental error sources, LOS or Non-Line of Sight (NLOS) conditions, can generally be modeled as Gaussian random variables~\cite{dong2021low}. In optimal LOS scenarios, the error can be as low as 20 centimeters~\cite{dong2021low}. Additionally, we use Decawave DWM1001 UWB modules, which offer accuracy of up to 5 centimeters. Data is collected using the ROS~2 with custom packages developed to interface with the modules.

\subsection{Data Acquisition}
Five volunteers actively participated in this study, with UWB nodes strategically placed on their ankles, wrists, and bellies to comprehensively capture a range of movements and postures. Engaging in nine distinct postures—standby, up, down, left, right, takeoff, land, forward, and backward—these individuals facilitated the collection of rich UWB data, providing a robust foundation for our model to accurately predict corresponding poses, which is particularly pertinent for guiding robots. The dataset presented in this work consists of these nine different human postures, utilizing Qorvo’s DWM1001 UWB modules with custom firmware for ToF ranging, all placed on the participants' bodies.


All measurements are obtained from the DWM1001 device via UART connection and interfaced with ROS~2 nodes for processing and recording. The five UWB nodes, positioned on the human body, communicate in an all-to-all manner, taking UWB measurements of the distances between nodes at a frequency of 15 Hz. For each posture, 400 measurements are recorded, ensuring a comprehensive dataset that captures the human movement and provides reliable data for model training and validation.

\subsection{Machine Learning algorithms}

Machine Learning aims to enable systems to learn from data, making predictions or identifying patterns~\cite{wu2024review}. Machine Learning methods are generally classified into supervised, unsupervised, and reinforcement learning~\cite{alpaydin2020introduction, wu2024review} where reinforcement learning optimizes actions through iterative interactions with the environment, guided by a reward system~\cite{sutton2018reinforcement}.

In this research, we collected UWB data from various human postures at a frequency of 15 Hz and employed machine learning and deep learning algorithms for human gesture prediction. Specifically, we compared the performance and accuracy of three classifiers: KNN, SVM, and MLP~\cite{krawczyk2023comparison, lin2023wi}, implemented using the scikit-learn (sklearn) library. The aim was to evaluate the efficiency of each method in accurately predicting human gestures and to identify which algorithm performs best under different conditions. This comparative analysis helps highlight the strengths and limitations of each approach within the proposed UWB-based posture estimation system, providing insights into their suitability for real-world applications. By understanding the performance variations among these models, we can better select the most appropriate algorithm for specific tasks, optimize computational resources, and improve the overall robustness and reliability of the human pose estimation system.

\subsubsection{KNN}

The KNN algorithm~\cite{kramer2013dimensionality} classifies data points based on the proximity of k-nearest neighbors, using the Manhattan distance for evaluation. Classification is determined by the majority label among these neighbors. KNN has been widely used in fingerprint identification techniques~\cite{yu20145,xie2016improved,hoang2018soft}, particularly for estimating the position of target devices. However, it requires maintaining large radio maps for frequent computations.

\subsubsection{SVM}

The SVM, a classical algorithm~\cite{suthaharan2016machine}, identifies the optimal hyperplane to separate data into classes, maximizing the margin between them. It is widely used in classification tasks, modeling both linear and non-linear relationships~\cite{sabanci2018wifi,yu2014indoor}. The kernel mechanism enhances its generalization, making SVM suitable for various fingerprint identification applications.

\subsubsection{MLP}

The MLP architecture, an artificial neural network~\cite{paegelow2018geomatic}, consists of an input layer, multiple hidden layers, and an output layer, with each layer containing neuron models. Neurons in adjacent layers are fully interconnected, enabling computation propagation from input to output during inference. Supervised training is performed using the backpropagation method, allowing MLP to transform input data into target results effectively. MLP's robust feature extraction capabilities make it suitable for various applications in neural networks and fingerprint analysis~\cite{zhu2020accurate}.

%% file: sec/03_ModelPerformance.tex

\section{Pose Classification}

\begin{figure*}[tb]
    \centering
    \includegraphics[width=1.0\textwidth]{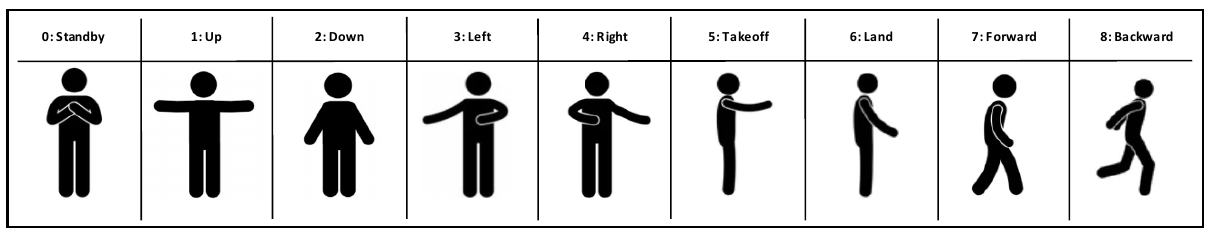}
    \caption{Showcase of the different classes}
    \label{fig:classes}
\end{figure*}

In this section, we present the classification performance of the different models introduced in the previous section. For each posture, 400 measurements were recorded from UWB nodes for five different participants. The data from four participants was used to train our models, while the data from the remaining participant was used to test our models. This approach ensures that the models are evaluated on unseen data, objectively assessing their predictive capabilities. For our posture prediction, we defined nine distinct classes from 0 to 8 as shown in Fig.~\ref{fig:classes}. This classification framework allows for precise identification and prediction of various human postures, which is essential for guiding robotic movements.

To further evaluate the performance and robustness of our models, we introduced noise ranging from 0 to 30 centimeters separately to the training data, the test data, or both. This testing scenario helps determine how well the models can handle real-world data imperfections and maintain accuracy under varying conditions.
For validating the models, we use a leave-one-out cross-validation methodology, where all subjects except one are used to train the model, and the remaining subject is used as the source of test data. This process is iterated over all subjects serving separately as test data sources. With this, we aim at demonstrating generalization as the models are not biased towards any particular individual's data while avoiding overfitting to the relatively limited amount of data, from the perspective of having different test subjects.


By thoroughly testing the models under these various conditions, we aim to identify the most robust and accurate approach for human gesture prediction, ensuring that the chosen model can perform effectively in diverse real-world scenarios.




\begin{figure*}[t]
    \centering
    \setlength\figureheight{0.3\textwidth}
    \setlength\figurewidth{0.95\textwidth}
    \scriptsize{\input{graphs/each_class_balanced_accuracy}}
    \caption{Each class balanced accuracy}
    \label{fig:each_class_balanced_accuracy}
\end{figure*}
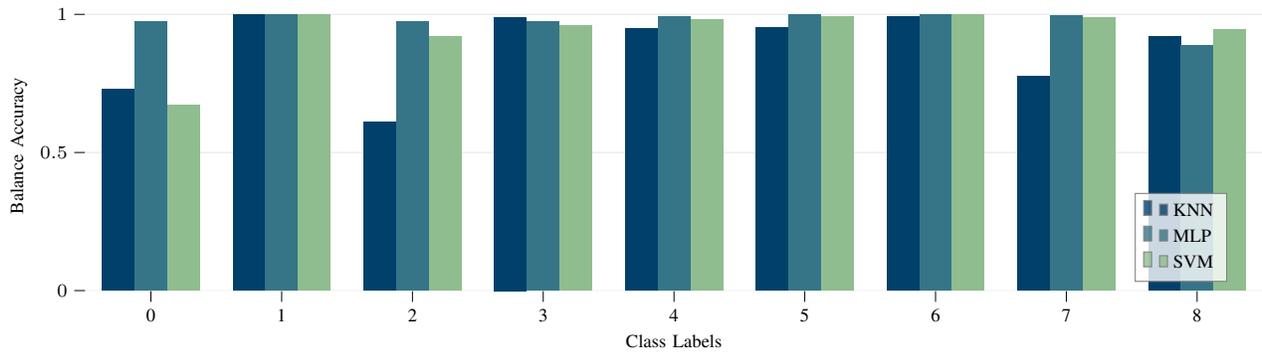

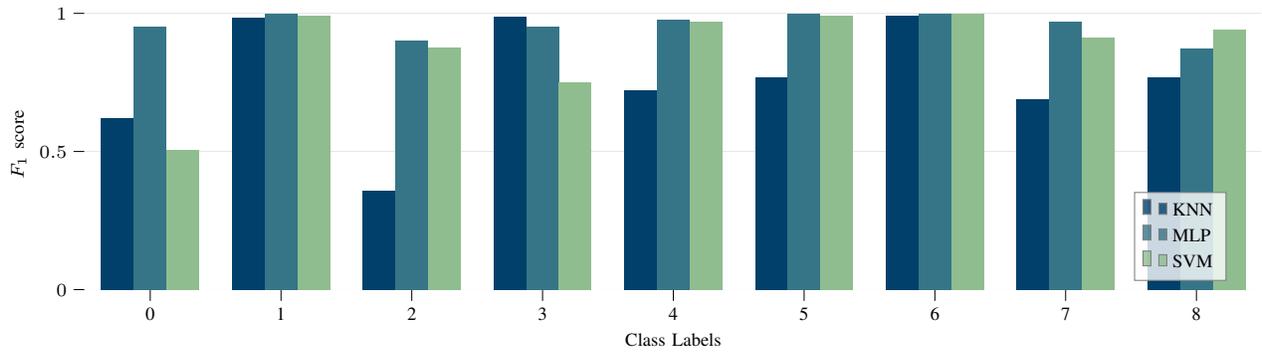
\begin{figure*}[t]
    \centering
    \setlength\figureheight{0.3\textwidth}
    \setlength\figurewidth{0.95\textwidth}
    \scriptsize{\input{graphs/each_class_f1_score}}
    \caption{Each class $F_{1}$ score}
    \label{fig:each_class_f1_score}
\end{figure*}

\subsubsection{KNN}

\begin{figure}[tb]
    \centering
    \setlength\figureheight{0.3\textwidth}
    \setlength\figurewidth{0.48\textwidth}
    \scriptsize{\input{graphs/knn_confusion_matrix}}
    \caption{KNN confusion matrix obtained with leave-one-out cross-validation (iterating over all subjects).}
    \label{fig:knn_confusion_matrix}
\end{figure}
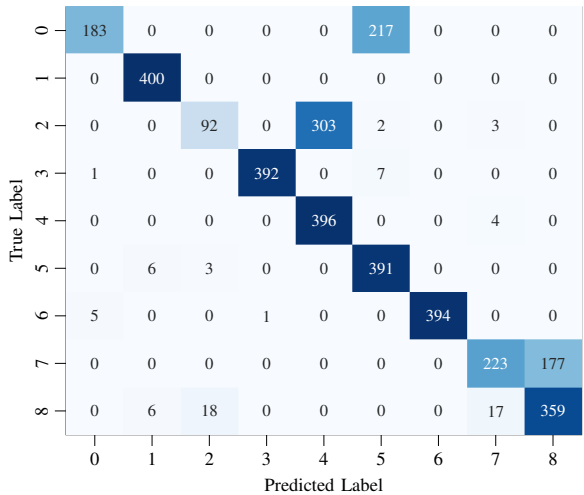

For the KNN model, the optimal value of k was determined using the elbow method. In the absence of noise, the model achieved an accuracy of 86\% for the test participant's data with the best k value being 2 where the confusion matrix can be seen in Fig.~\ref{fig:knn_confusion_matrix}. Detailed balanced accuracy and $F_{1}$ score for each class are illustrated in Fig.~\ref{fig:each_class_balanced_accuracy} and Fig.~\ref{fig:each_class_f1_score}, respectively. As shown, the model achieved a balanced accuracy of 73\% for class 0, indicating moderate performance. However, the model struggled with class 2, achieving only a 61\% balanced accuracy. These results are mirrored in the F1 scores, where class 0 has an F1 score of 62\%, and class 2 shows an even lower F1 score of 36\%. The model performed exceptionally well for other classes.

When noise was introduced, ranging from 0 to 30 centimeters, the model’s performance varied. With noise added to the training data, the accuracy dropped to 79\%, and the optimal k value increased to 4. Adding noise solely to the test data resulted in an accuracy of 75\%, while keeping the best k value at 2. Introducing noise to both training and test data led to an accuracy of 76\%, with the optimal k value remaining at 2. These results indicate that while KNN is sensitive to noise, it retains reasonable performance with appropriate adjustments to the k value across different scenarios.

\subsubsection{SVM}

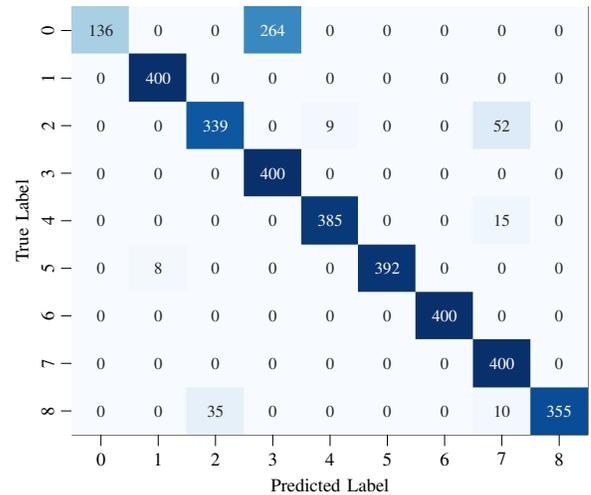
\begin{figure}[tb]
    \centering
    \setlength\figureheight{0.3\textwidth}
    \setlength\figurewidth{0.48\textwidth}
    \scriptsize{\input{graphs/svm_confusion_matrix}}
    \caption{SVM confusion matrix obtained with leave-one-out cross-validation (iterating over all subjects).}
    \label{fig:svm_confusion_matrix}
\end{figure}

For the SVM model, Gridsearch was used to find the optimal parameters, such as C and gamma values. In the absence of noise, the model achieved an accuracy of 97\% with C equal to 1 and gamma equal to 0.1 where the confusion matrix can be seen in Fig.~\ref{fig:svm_confusion_matrix}. As it can be seen in Fig.~\ref{fig:each_class_balanced_accuracy}, SVM model exhibited robust performance across all classes, maintaining high balanced accuracy and F1 scores. The model achieved a perfect balanced accuracy of 100\% for class 6 and 1, reflecting flawless performance, and maintained performance above 95\% for most other classes. The F1 scores as shown in Fig.~\ref{fig:each_class_f1_score} corroborate these findings, with a perfect score for classes 6 and 2 and high scores for other classes, including 97\% for class 4 and 99\% for class 5.

When noise levels between 0 and 30 centimeters were applied, the model's accuracy decreased to 86\% when noise was added to the training data, with the optimal C value shifting to 10 while gamma remained at 0.1. Introducing noise only in the test data reduced the accuracy to 81\%, maintaining C at 1 and gamma at 0.1. When noise was present in both the training and test data, the accuracy improved to 92\%, with C set to 10 and gamma to 0.1. These results underscore the SVM model's robustness and adaptability to noise, maintaining high accuracy across varying conditions.

\subsubsection{MLP}

\begin{figure}[tb]
    \centering
    \setlength\figureheight{0.3\textwidth}
    \setlength\figurewidth{0.48\textwidth}
    \scriptsize{\input{graphs/mlp_confusion_matrix}}
    \caption{MLP confusion matrix obtained with leave-one-out cross-validation (iterating over all subjects).}
    \label{fig:mlp_confusion_matrix}
\end{figure}
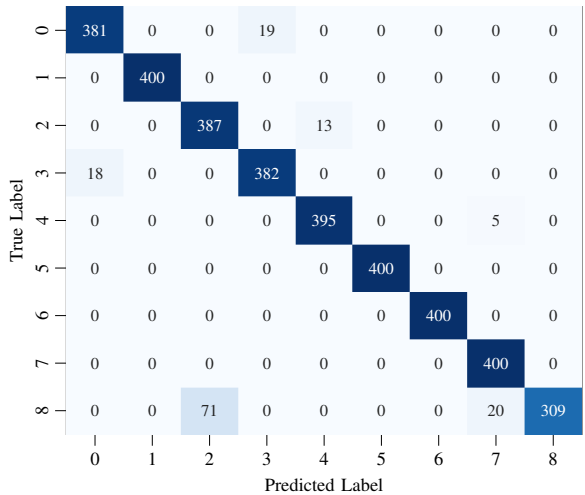

For the MLP model, the accuracy was 96\% in the absence of noise where the confusion matrix can be seen in Fig.~\ref{fig:mlp_confusion_matrix}. As can be seen in Fig.~\ref{fig:each_class_balanced_accuracy} and Fig.~\ref{fig:each_class_f1_score}, MLP model demonstrated exceptional performance in human posture estimation, achieving high balanced accuracy and F1 scores across most classes. The model achieved perfect balanced accuracy for classes 1, 5, 6 and 7, reflecting its strong classification capabilities. Corresponding F1 scores were also perfect for classes 1,5 and 6, with additional high scores of 98\% for class 4 and 97\% for class 7. While the model showed slight performance dips for classes 2 and 8, the balanced accuracies obtained were 97\% and 89\%, respectively, and F1 scores of 90\% and 87\%.

With the introduction of noise between 0 and 30 centimeters, the model's accuracy dropped to 83\% when noise was added to the training data. Noise only in the test data reduced accuracy to 80\%. Finally, when noise was introduced to both training and test data, the accuracy improved to 89\%. These results highlight the MLP model's robustness, showing it can maintain strong performance even in the presence of noise, making it a reliable choice for precise human posture estimation tasks.

\begin{figure*}[tb]
    \centering
    \setlength\figureheight{0.3\textwidth}
    \setlength\figurewidth{0.48\textwidth}
    \scriptsize{\input{graphs/main_table_balanced_accuracy}}
    \caption{Assessment of the maximum and minimum counts of UWB nodes}
    \label{fig:main_table}
\end{figure*}
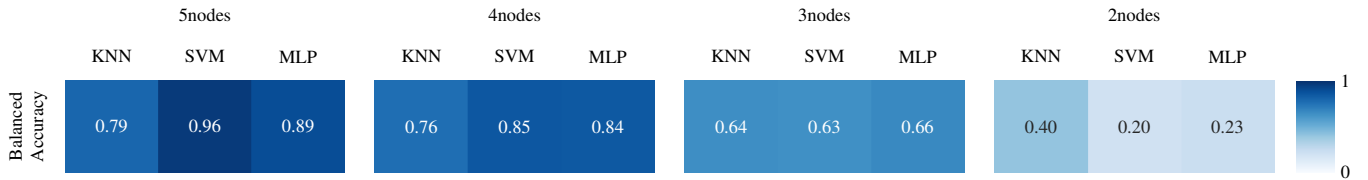

In this study, we used five UWB nodes for data recording and model training to predict human postures accurately and send movement commands to robots. To determine the minimum number of nodes required for accurate posture prediction, we tested the same procedure with different numbers of UWB nodes, as the average balanced accuracy illustrated in Fig.~\ref{fig:main_table}. The results demonstrate a clear trend: the reduction in the number of nodes generally leads to a decrease in balanced accuracy across all models. With five nodes, the MLP model achieved the highest balanced accuracy at 0.89, followed by SVM at 0.96 and KNN at 0.79. As the number of nodes decreased to four, all models experienced a slight drop in performance, with MLP and SVM still maintaining high balanced accuracy values of 0.84 and 0.85, respectively, while KNN dropped to 0.76. Further reductions to three nodes resulted in a more noticeable decline in accuracy for all models indicating that three nodes might be nearing the lower limit for effective posture prediction. When reduced to two nodes, the performance drop was significant suggesting that two nodes are insufficient for reliable posture estimation.

These findings highlight the importance of the number of UWB nodes in maintaining high prediction accuracy. While the MLP and SVM models showed some resilience to the reduction in nodes, the overall trend indicates that a minimum of four nodes is necessary to achieve reasonably accurate posture predictions. This reduction in node count can streamline the setup and simplify deployment while still ensuring reliable performance for robotic movement commands.

%% file: graphs/each_class_balanced_accuracy.tex
\begin{tikzpicture}

\definecolor{darkgray176}{RGB}{176,176,176}
\definecolor{green}{RGB}{0,128,0}
\definecolor{orange}{RGB}{255,165,0}
\definecolor{skyblue}{RGB}{135,206,235}

\definecolor{color1}{rgb}{0.0, 0.25, 0.42}
\definecolor{color2}{rgb}{0.21, 0.46, 0.53}
\definecolor{color3}{rgb}{0.56, 0.74, 0.56}    

\begin{axis}[
    width=\figurewidth,
    height=\figureheight,
    axis line style={white},
    legend cell align={left},
    legend style={
      fill opacity=0.8,
      draw opacity=1,
      text opacity=1,
      at={(0.03,0.97)},
      anchor=north west,
      draw=gray,
    },
    legend pos=south east,
    tick align=outside,
    tick pos=left,
    x grid style={white!90!black},
    xlabel={Class Labels},
    xmin=-0.5, xmax=8.5,
    xtick={0,1,2,3,4,5,6,7,8},
    xticklabels={0, 1,2,3,4,5,6,7,8},
    xtick style={color=black},
    y grid style={white!90!black},
    ylabel={Balance Accuracy},
    ymajorgrids,
    ymin=0, ymax=1.05,
    yminorgrids,
    ytick style={color=black}
]

\draw[draw=none,fill=color1] (axis cs:-0.375,0) rectangle (axis cs:-0.125,0.7278125);
\draw[draw=none,fill=color1] (axis cs:0.625,0) rectangle (axis cs:0.875,0.9981249999999999);
\draw[draw=none,fill=color1] (axis cs:1.625,0) rectangle (axis cs:1.875,0.61171875);
\draw[draw=none,fill=color1] (axis cs:2.625,0) rectangle (axis cs:2.875,0.9898437499999999);
\draw[draw=none,fill=color1] (axis cs:3.625,0) rectangle (axis cs:3.875,0.94765625);
\draw[draw=none,fill=color1] (axis cs:4.625,0) rectangle (axis cs:4.875,0.9534374999999999);
\draw[draw=none,fill=color1] (axis cs:5.625,0) rectangle (axis cs:5.875,0.9924999999999999);
\draw[draw=none,fill=color1] (axis cs:6.625,0) rectangle (axis cs:6.875,0.775);
\draw[draw=none,fill=color1] (axis cs:7.625,0) rectangle (axis cs:7.875,0.92109375);

\addlegendimage{ybar,ybar legend,draw=none,fill=color1}
\addlegendentry{KNN}

\draw[draw=none,fill=color2] (axis cs:-0.125,0) rectangle (axis cs:0.125,0.9734375);
\draw[draw=none,fill=color2] (axis cs:0.875,0) rectangle (axis cs:1.125,1.0);
\draw[draw=none,fill=color2] (axis cs:1.875,0) rectangle (axis cs:2.125,0.97265625);
\draw[draw=none,fill=color2] (axis cs:2.875,0) rectangle (axis cs:3.125,0.97453125);
\draw[draw=none,fill=color2] (axis cs:3.875,0) rectangle (axis cs:4.125,0.99171875);
\draw[draw=none,fill=color2] (axis cs:4.875,0) rectangle (axis cs:5.125,1.0);
\draw[draw=none,fill=color2] (axis cs:5.875,0) rectangle (axis cs:6.125,1.0);
\draw[draw=none,fill=color2] (axis cs:6.875,0) rectangle (axis cs:7.125,0.99609375);
\draw[draw=none,fill=color2] (axis cs:7.875,0) rectangle (axis cs:8.125,0.88625);

\addlegendimage{ybar,ybar legend,draw=none,fill=color2}
\addlegendentry{MLP}

\draw[draw=none,fill=color3] (axis cs:0.125,0) rectangle (axis cs:0.375,0.67);
\draw[draw=none,fill=color3] (axis cs:1.125,0) rectangle (axis cs:1.375,0.99875);
\draw[draw=none,fill=color3] (axis cs:2.125,0) rectangle (axis cs:2.375,0.9182812499999999);
\draw[draw=none,fill=color3] (axis cs:3.125,0) rectangle (axis cs:3.375,0.95875);
\draw[draw=none,fill=color3] (axis cs:4.125,0) rectangle (axis cs:4.375,0.97984375);
\draw[draw=none,fill=color3] (axis cs:5.125,0) rectangle (axis cs:5.375,0.99);
\draw[draw=none,fill=color3] (axis cs:6.125,0) rectangle (axis cs:6.375,1.0);
\draw[draw=none,fill=color3] (axis cs:7.125,0) rectangle (axis cs:7.375,0.9879687500000001);
\draw[draw=none,fill=color3] (axis cs:8.125,0) rectangle (axis cs:8.375,0.94375);

\addlegendimage{ybar,ybar legend,draw=none,fill=color3}
\addlegendentry{SVM}
\end{axis}

\end{tikzpicture}

%% file: graphs/each_class_f1_score.tex
\begin{tikzpicture}

\definecolor{darkgray176}{RGB}{176,176,176}
\definecolor{green}{RGB}{0,128,0}
\definecolor{orange}{RGB}{255,165,0}
\definecolor{skyblue}{RGB}{135,206,235}

\definecolor{color1}{rgb}{0.0, 0.25, 0.42}
\definecolor{color2}{rgb}{0.21, 0.46, 0.53}
\definecolor{color3}{rgb}{0.56, 0.74, 0.56}    

\begin{axis}[
    width=\figurewidth,
    height=\figureheight,
    axis line style={white},
    legend cell align={left},
    legend style={
      fill opacity=0.8,
      draw opacity=1,
      text opacity=1,
      at={(0.03,0.97)},
      anchor=north west,
      draw=gray,
    },
    legend pos=south east,
    tick align=outside,
    tick pos=left,
    x grid style={white!90!black},
    xlabel={Class Labels},
    xmin=-0.5, xmax=8.5,
    xtick={0,1,2,3,4,5,6,7,8},
    xticklabels={0, 1,2,3,4,5,6,7,8},
    xtick style={color=black},
    y grid style={white!90!black},
    ylabel={$F_{1}$ score},
    ymajorgrids,
    ymin=0, ymax=1.05,
    yminorgrids,
    ytick style={color=black}
]

\draw[draw=none,fill=color1] (axis cs:-0.375,0) rectangle (axis cs:-0.125,0.6213921901528013);
\draw[draw=none,fill=color1] (axis cs:0.625,0) rectangle (axis cs:0.875,0.9852216748768473);
\draw[draw=none,fill=color1] (axis cs:1.625,0) rectangle (axis cs:1.875,0.3586744639376218);
\draw[draw=none,fill=color1] (axis cs:2.625,0) rectangle (axis cs:2.875,0.9886506935687264);
\draw[draw=none,fill=color1] (axis cs:3.625,0) rectangle (axis cs:3.875,0.7206551410373065);
\draw[draw=none,fill=color1] (axis cs:4.625,0) rectangle (axis cs:4.875,0.768928220255654);
\draw[draw=none,fill=color1] (axis cs:5.625,0) rectangle (axis cs:5.875,0.9924433249370278);
\draw[draw=none,fill=color1] (axis cs:6.625,0) rectangle (axis cs:6.875,0.6893353941267387);
\draw[draw=none,fill=color1] (axis cs:7.625,0) rectangle (axis cs:7.875,0.7670940170940171);

\addlegendimage{ybar,ybar legend,draw=none,fill=color1}
\addlegendentry{KNN}

\draw[draw=none,fill=color2] (axis cs:-0.125,0) rectangle (axis cs:0.125,0.9536921151439299);
\draw[draw=none,fill=color2] (axis cs:0.875,0) rectangle (axis cs:1.125,1.0);
\draw[draw=none,fill=color2] (axis cs:1.875,0) rectangle (axis cs:2.125,0.9020979020979021);
\draw[draw=none,fill=color2] (axis cs:2.875,0) rectangle (axis cs:3.125,0.9538077403245941);
\draw[draw=none,fill=color2] (axis cs:3.875,0) rectangle (axis cs:4.125,0.9777227722772278);
\draw[draw=none,fill=color2] (axis cs:4.875,0) rectangle (axis cs:5.125,1.0);
\draw[draw=none,fill=color2] (axis cs:5.875,0) rectangle (axis cs:6.125,1.0);
\draw[draw=none,fill=color2] (axis cs:6.875,0) rectangle (axis cs:7.125,0.9696969696969697);
\draw[draw=none,fill=color2] (axis cs:7.875,0) rectangle (axis cs:8.125,0.8716502115655853);

\addlegendimage{ybar,ybar legend,draw=none,fill=color2}
\addlegendentry{MLP}

\draw[draw=none,fill=color3] (axis cs:0.125,0) rectangle (axis cs:0.375,0.5074626865671642);
\draw[draw=none,fill=color3] (axis cs:1.125,0) rectangle (axis cs:1.375,0.99009900990099);
\draw[draw=none,fill=color3] (axis cs:2.125,0) rectangle (axis cs:2.375, 0.875968992248062);
\draw[draw=none,fill=color3] (axis cs:3.125,0) rectangle (axis cs:3.375,0.7518796992481204);
\draw[draw=none,fill=color3] (axis cs:4.125,0) rectangle (axis cs:4.375,0.9697732997481108);
\draw[draw=none,fill=color3] (axis cs:5.125,0) rectangle (axis cs:5.375,0.98989898989899);
\draw[draw=none,fill=color3] (axis cs:6.125,0) rectangle (axis cs:6.375,1.0);
\draw[draw=none,fill=color3] (axis cs:7.125,0) rectangle (axis cs:7.375,0.9122006841505131);
\draw[draw=none,fill=color3] (axis cs:8.125,0) rectangle (axis cs:8.375,0.9403973509933775);

\addlegendimage{ybar,ybar legend,draw=none,fill=color3}
\addlegendentry{SVM}

\end{axis}

\end{tikzpicture}

%% file: graphs/knn_confusion_matrix.tex
\begin{tikzpicture}

\definecolor{darkgray176}{RGB}{176,176,176}
\definecolor{darkslategray38}{RGB}{38,38,38}

\begin{axis}[
tick align=outside,
tick pos=left,
x grid style={darkgray176},
xlabel={Predicted Label},
xmin=0, xmax=9,
xtick style={color=black},
xtick={0.5,1.5,2.5,3.5,4.5,5.5,6.5,7.5,8.5},
xticklabels={0,1,2,3,4,5,6,7,8},
y dir=reverse,
y grid style={darkgray176},
ylabel={True Label},
ymin=0, ymax=9,
ytick style={color=black},
ytick={0.5,1.5,2.5,3.5,4.5,5.5,6.5,7.5,8.5},
yticklabel style={rotate=90.0},
yticklabels={0,1,2,3,4,5,6,7,8}
]
\addplot graphics [includegraphics cmd=\pgfimage,xmin=0, xmax=9, ymin=9, ymax=0] {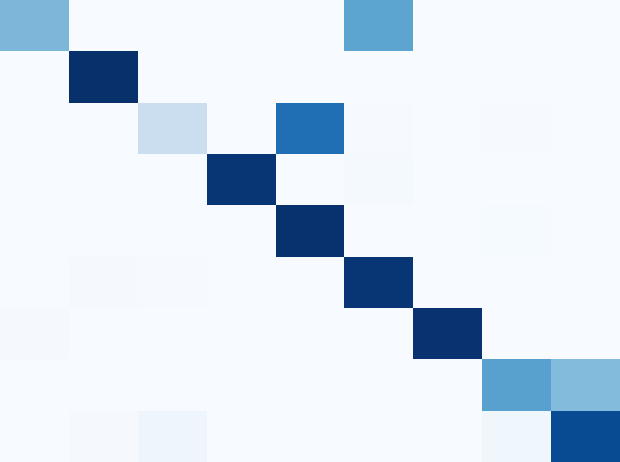};
\draw (axis cs:0.5,0.5) node[
  scale=0.9,
  text=darkslategray38,
  rotate=0.0
]{183};
\draw (axis cs:1.5,0.5) node[
  scale=0.9,
  text=darkslategray38,
  rotate=0.0
]{0};
\draw (axis cs:2.5,0.5) node[
  scale=0.9,
  text=darkslategray38,
  rotate=0.0
]{0};
\draw (axis cs:3.5,0.5) node[
  scale=0.9,
  text=darkslategray38,
  rotate=0.0
]{0};
\draw (axis cs:4.5,0.5) node[
  scale=0.9,
  text=darkslategray38,
  rotate=0.0
]{0};
\draw (axis cs:5.5,0.5) node[
  scale=0.9,
  text=white,
  rotate=0.0
]{217};
\draw (axis cs:6.5,0.5) node[
  scale=0.9,
  text=darkslategray38,
  rotate=0.0
]{0};
\draw (axis cs:7.5,0.5) node[
  scale=0.9,
  text=darkslategray38,
  rotate=0.0
]{0};
\draw (axis cs:8.5,0.5) node[
  scale=0.9,
  text=darkslategray38,
  rotate=0.0
]{0};
\draw (axis cs:0.5,1.5) node[
  scale=0.9,
  text=darkslategray38,
  rotate=0.0
]{0};
\draw (axis cs:1.5,1.5) node[
  scale=0.9,
  text=white,
  rotate=0.0
]{400};
\draw (axis cs:2.5,1.5) node[
  scale=0.9,
  text=darkslategray38,
  rotate=0.0
]{0};
\draw (axis cs:3.5,1.5) node[
  scale=0.9,
  text=darkslategray38,
  rotate=0.0
]{0};
\draw (axis cs:4.5,1.5) node[
  scale=0.9,
  text=darkslategray38,
  rotate=0.0
]{0};
\draw (axis cs:5.5,1.5) node[
  scale=0.9,
  text=darkslategray38,
  rotate=0.0
]{0};
\draw (axis cs:6.5,1.5) node[
  scale=0.9,
  text=darkslategray38,
  rotate=0.0
]{0};
\draw (axis cs:7.5,1.5) node[
  scale=0.9,
  text=darkslategray38,
  rotate=0.0
]{0};
\draw (axis cs:8.5,1.5) node[
  scale=0.9,
  text=darkslategray38,
  rotate=0.0
]{0};
\draw (axis cs:0.5,2.5) node[
  scale=0.9,
  text=darkslategray38,
  rotate=0.0
]{0};
\draw (axis cs:1.5,2.5) node[
  scale=0.9,
  text=darkslategray38,
  rotate=0.0
]{0};
\draw (axis cs:2.5,2.5) node[
  scale=0.9,
  text=darkslategray38,
  rotate=0.0
]{92};
\draw (axis cs:3.5,2.5) node[
  scale=0.9,
  text=darkslategray38,
  rotate=0.0
]{0};
\draw (axis cs:4.5,2.5) node[
  scale=0.9,
  text=white,
  rotate=0.0
]{303};
\draw (axis cs:5.5,2.5) node[
  scale=0.9,
  text=darkslategray38,
  rotate=0.0
]{2};
\draw (axis cs:6.5,2.5) node[
  scale=0.9,
  text=darkslategray38,
  rotate=0.0
]{0};
\draw (axis cs:7.5,2.5) node[
  scale=0.9,
  text=darkslategray38,
  rotate=0.0
]{3};
\draw (axis cs:8.5,2.5) node[
  scale=0.9,
  text=darkslategray38,
  rotate=0.0
]{0};
\draw (axis cs:0.5,3.5) node[
  scale=0.9,
  text=darkslategray38,
  rotate=0.0
]{1};
\draw (axis cs:1.5,3.5) node[
  scale=0.9,
  text=darkslategray38,
  rotate=0.0
]{0};
\draw (axis cs:2.5,3.5) node[
  scale=0.9,
  text=darkslategray38,
  rotate=0.0
]{0};
\draw (axis cs:3.5,3.5) node[
  scale=0.9,
  text=white,
  rotate=0.0
]{392};
\draw (axis cs:4.5,3.5) node[
  scale=0.9,
  text=darkslategray38,
  rotate=0.0
]{0};
\draw (axis cs:5.5,3.5) node[
  scale=0.9,
  text=darkslategray38,
  rotate=0.0
]{7};
\draw (axis cs:6.5,3.5) node[
  scale=0.9,
  text=darkslategray38,
  rotate=0.0
]{0};
\draw (axis cs:7.5,3.5) node[
  scale=0.9,
  text=darkslategray38,
  rotate=0.0
]{0};
\draw (axis cs:8.5,3.5) node[
  scale=0.9,
  text=darkslategray38,
  rotate=0.0
]{0};
\draw (axis cs:0.5,4.5) node[
  scale=0.9,
  text=darkslategray38,
  rotate=0.0
]{0};
\draw (axis cs:1.5,4.5) node[
  scale=0.9,
  text=darkslategray38,
  rotate=0.0
]{0};
\draw (axis cs:2.5,4.5) node[
  scale=0.9,
  text=darkslategray38,
  rotate=0.0
]{0};
\draw (axis cs:3.5,4.5) node[
  scale=0.9,
  text=darkslategray38,
  rotate=0.0
]{0};
\draw (axis cs:4.5,4.5) node[
  scale=0.9,
  text=white,
  rotate=0.0
]{396};
\draw (axis cs:5.5,4.5) node[
  scale=0.9,
  text=darkslategray38,
  rotate=0.0
]{0};
\draw (axis cs:6.5,4.5) node[
  scale=0.9,
  text=darkslategray38,
  rotate=0.0
]{0};
\draw (axis cs:7.5,4.5) node[
  scale=0.9,
  text=darkslategray38,
  rotate=0.0
]{4};
\draw (axis cs:8.5,4.5) node[
  scale=0.9,
  text=darkslategray38,
  rotate=0.0
]{0};
\draw (axis cs:0.5,5.5) node[
  scale=0.9,
  text=darkslategray38,
  rotate=0.0
]{0};
\draw (axis cs:1.5,5.5) node[
  scale=0.9,
  text=darkslategray38,
  rotate=0.0
]{6};
\draw (axis cs:2.5,5.5) node[
  scale=0.9,
  text=darkslategray38,
  rotate=0.0
]{3};
\draw (axis cs:3.5,5.5) node[
  scale=0.9,
  text=darkslategray38,
  rotate=0.0
]{0};
\draw (axis cs:4.5,5.5) node[
  scale=0.9,
  text=darkslategray38,
  rotate=0.0
]{0};
\draw (axis cs:5.5,5.5) node[
  scale=0.9,
  text=white,
  rotate=0.0
]{391};
\draw (axis cs:6.5,5.5) node[
  scale=0.9,
  text=darkslategray38,
  rotate=0.0
]{0};
\draw (axis cs:7.5,5.5) node[
  scale=0.9,
  text=darkslategray38,
  rotate=0.0
]{0};
\draw (axis cs:8.5,5.5) node[
  scale=0.9,
  text=darkslategray38,
  rotate=0.0
]{0};
\draw (axis cs:0.5,6.5) node[
  scale=0.9,
  text=darkslategray38,
  rotate=0.0
]{5};
\draw (axis cs:1.5,6.5) node[
  scale=0.9,
  text=darkslategray38,
  rotate=0.0
]{0};
\draw (axis cs:2.5,6.5) node[
  scale=0.9,
  text=darkslategray38,
  rotate=0.0
]{0};
\draw (axis cs:3.5,6.5) node[
  scale=0.9,
  text=darkslategray38,
  rotate=0.0
]{1};
\draw (axis cs:4.5,6.5) node[
  scale=0.9,
  text=darkslategray38,
  rotate=0.0
]{0};
\draw (axis cs:5.5,6.5) node[
  scale=0.9,
  text=darkslategray38,
  rotate=0.0
]{0};
\draw (axis cs:6.5,6.5) node[
  scale=0.9,
  text=white,
  rotate=0.0
]{394};
\draw (axis cs:7.5,6.5) node[
  scale=0.9,
  text=darkslategray38,
  rotate=0.0
]{0};
\draw (axis cs:8.5,6.5) node[
  scale=0.9,
  text=darkslategray38,
  rotate=0.0
]{0};
\draw (axis cs:0.5,7.5) node[
  scale=0.9,
  text=darkslategray38,
  rotate=0.0
]{0};
\draw (axis cs:1.5,7.5) node[
  scale=0.9,
  text=darkslategray38,
  rotate=0.0
]{0};
\draw (axis cs:2.5,7.5) node[
  scale=0.9,
  text=darkslategray38,
  rotate=0.0
]{0};
\draw (axis cs:3.5,7.5) node[
  scale=0.9,
  text=darkslategray38,
  rotate=0.0
]{0};
\draw (axis cs:4.5,7.5) node[
  scale=0.9,
  text=darkslategray38,
  rotate=0.0
]{0};
\draw (axis cs:5.5,7.5) node[
  scale=0.9,
  text=darkslategray38,
  rotate=0.0
]{0};
\draw (axis cs:6.5,7.5) node[
  scale=0.9,
  text=darkslategray38,
  rotate=0.0
]{0};
\draw (axis cs:7.5,7.5) node[
  scale=0.9,
  text=white,
  rotate=0.0
]{223};
\draw (axis cs:8.5,7.5) node[
  scale=0.9,
  text=darkslategray38,
  rotate=0.0
]{177};
\draw (axis cs:0.5,8.5) node[
  scale=0.9,
  text=darkslategray38,
  rotate=0.0
]{0};
\draw (axis cs:1.5,8.5) node[
  scale=0.9,
  text=darkslategray38,
  rotate=0.0
]{6};
\draw (axis cs:2.5,8.5) node[
  scale=0.9,
  text=darkslategray38,
  rotate=0.0
]{18};
\draw (axis cs:3.5,8.5) node[
  scale=0.9,
  text=darkslategray38,
  rotate=0.0
]{0};
\draw (axis cs:4.5,8.5) node[
  scale=0.9,
  text=darkslategray38,
  rotate=0.0
]{0};
\draw (axis cs:5.5,8.5) node[
  scale=0.9,
  text=darkslategray38,
  rotate=0.0
]{0};
\draw (axis cs:6.5,8.5) node[
  scale=0.9,
  text=darkslategray38,
  rotate=0.0
]{0};
\draw (axis cs:7.5,8.5) node[
  scale=0.9,
  text=darkslategray38,
  rotate=0.0
]{17};
\draw (axis cs:8.5,8.5) node[
  scale=0.9,
  text=white,
  rotate=0.0
]{359};
\end{axis}

\end{tikzpicture}

%% file: graphs/svm_confusion_matrix.tex
\begin{tikzpicture}

  \definecolor{darkgray176}{RGB}{176,176,176}
  \definecolor{darkslategray38}{RGB}{38,38,38}
  
  \begin{axis}[
  tick align=outside,
  tick pos=left,
  x grid style={darkgray176},
  xlabel={Predicted Label},
  xmin=0, xmax=9,
  xtick style={color=black},
  xtick={0.5,1.5,2.5,3.5,4.5,5.5,6.5,7.5,8.5},
  xticklabels={0,1,2,3,4,5,6,7,8},
  y dir=reverse,
  y grid style={darkgray176},
  ylabel={True Label},
  ymin=0, ymax=9,
  ytick style={color=black},
  ytick={0.5,1.5,2.5,3.5,4.5,5.5,6.5,7.5,8.5},
  yticklabel style={rotate=90.0},
  yticklabels={0,1,2,3,4,5,6,7,8}
  ]
  \addplot graphics [includegraphics cmd=\pgfimage,xmin=0, xmax=9, ymin=9, ymax=0] {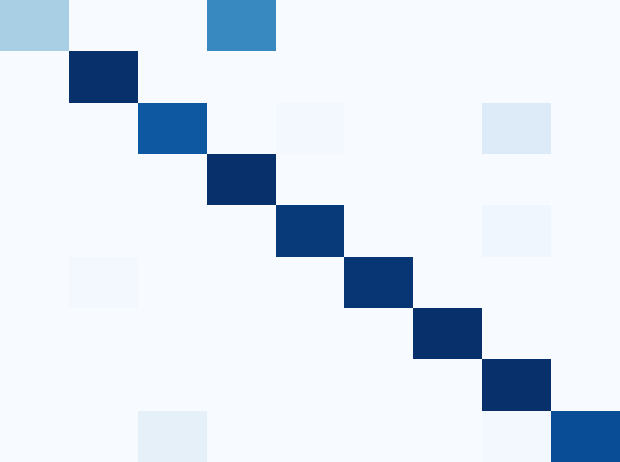};
  \draw (axis cs:0.5,0.5) node[
    scale=0.9,
    text=darkslategray38,
    rotate=0.0
  ]{136};
  \draw (axis cs:1.5,0.5) node[
    scale=0.9,
    text=darkslategray38,
    rotate=0.0
  ]{0};
  \draw (axis cs:2.5,0.5) node[
    scale=0.9,
    text=darkslategray38,
    rotate=0.0
  ]{0};
  \draw (axis cs:3.5,0.5) node[
    scale=0.9,
    text=white,
    rotate=0.0
  ]{264};
  \draw (axis cs:4.5,0.5) node[
    scale=0.9,
    text=darkslategray38,
    rotate=0.0
  ]{0};
  \draw (axis cs:5.5,0.5) node[
    scale=0.9,
    text=darkslategray38,
    rotate=0.0
  ]{0};
  \draw (axis cs:6.5,0.5) node[
    scale=0.9,
    text=darkslategray38,
    rotate=0.0
  ]{0};
  \draw (axis cs:7.5,0.5) node[
    scale=0.9,
    text=darkslategray38,
    rotate=0.0
  ]{0};
  \draw (axis cs:8.5,0.5) node[
    scale=0.9,
    text=darkslategray38,
    rotate=0.0
  ]{0};
  \draw (axis cs:0.5,1.5) node[
    scale=0.9,
    text=darkslategray38,
    rotate=0.0
  ]{0};
  \draw (axis cs:1.5,1.5) node[
    scale=0.9,
    text=white,
    rotate=0.0
  ]{400};
  \draw (axis cs:2.5,1.5) node[
    scale=0.9,
    text=darkslategray38,
    rotate=0.0
  ]{0};
  \draw (axis cs:3.5,1.5) node[
    scale=0.9,
    text=darkslategray38,
    rotate=0.0
  ]{0};
  \draw (axis cs:4.5,1.5) node[
    scale=0.9,
    text=darkslategray38,
    rotate=0.0
  ]{0};
  \draw (axis cs:5.5,1.5) node[
    scale=0.9,
    text=darkslategray38,
    rotate=0.0
  ]{0};
  \draw (axis cs:6.5,1.5) node[
    scale=0.9,
    text=darkslategray38,
    rotate=0.0
  ]{0};
  \draw (axis cs:7.5,1.5) node[
    scale=0.9,
    text=darkslategray38,
    rotate=0.0
  ]{0};
  \draw (axis cs:8.5,1.5) node[
    scale=0.9,
    text=darkslategray38,
    rotate=0.0
  ]{0};
  \draw (axis cs:0.5,2.5) node[
    scale=0.9,
    text=darkslategray38,
    rotate=0.0
  ]{0};
  \draw (axis cs:1.5,2.5) node[
    scale=0.9,
    text=darkslategray38,
    rotate=0.0
  ]{0};
  \draw (axis cs:2.5,2.5) node[
    scale=0.9,
    text=white,
    rotate=0.0
  ]{339};
  \draw (axis cs:3.5,2.5) node[
    scale=0.9,
    text=darkslategray38,
    rotate=0.0
  ]{0};
  \draw (axis cs:4.5,2.5) node[
    scale=0.9,
    text=darkslategray38,
    rotate=0.0
  ]{9};
  \draw (axis cs:5.5,2.5) node[
    scale=0.9,
    text=darkslategray38,
    rotate=0.0
  ]{0};
  \draw (axis cs:6.5,2.5) node[
    scale=0.9,
    text=darkslategray38,
    rotate=0.0
  ]{0};
  \draw (axis cs:7.5,2.5) node[
    scale=0.9,
    text=darkslategray38,
    rotate=0.0
  ]{52};
  \draw (axis cs:8.5,2.5) node[
    scale=0.9,
    text=darkslategray38,
    rotate=0.0
  ]{0};
  \draw (axis cs:0.5,3.5) node[
    scale=0.9,
    text=darkslategray38,
    rotate=0.0
  ]{0};
  \draw (axis cs:1.5,3.5) node[
    scale=0.9,
    text=darkslategray38,
    rotate=0.0
  ]{0};
  \draw (axis cs:2.5,3.5) node[
    scale=0.9,
    text=darkslategray38,
    rotate=0.0
  ]{0};
  \draw (axis cs:3.5,3.5) node[
    scale=0.9,
    text=white,
    rotate=0.0
  ]{400};
  \draw (axis cs:4.5,3.5) node[
    scale=0.9,
    text=darkslategray38,
    rotate=0.0
  ]{0};
  \draw (axis cs:5.5,3.5) node[
    scale=0.9,
    text=darkslategray38,
    rotate=0.0
  ]{0};
  \draw (axis cs:6.5,3.5) node[
    scale=0.9,
    text=darkslategray38,
    rotate=0.0
  ]{0};
  \draw (axis cs:7.5,3.5) node[
    scale=0.9,
    text=darkslategray38,
    rotate=0.0
  ]{0};
  \draw (axis cs:8.5,3.5) node[
    scale=0.9,
    text=darkslategray38,
    rotate=0.0
  ]{0};
  \draw (axis cs:0.5,4.5) node[
    scale=0.9,
    text=darkslategray38,
    rotate=0.0
  ]{0};
  \draw (axis cs:1.5,4.5) node[
    scale=0.9,
    text=darkslategray38,
    rotate=0.0
  ]{0};
  \draw (axis cs:2.5,4.5) node[
    scale=0.9,
    text=darkslategray38,
    rotate=0.0
  ]{0};
  \draw (axis cs:3.5,4.5) node[
    scale=0.9,
    text=darkslategray38,
    rotate=0.0
  ]{0};
  \draw (axis cs:4.5,4.5) node[
    scale=0.9,
    text=white,
    rotate=0.0
  ]{385};
  \draw (axis cs:5.5,4.5) node[
    scale=0.9,
    text=darkslategray38,
    rotate=0.0
  ]{0};
  \draw (axis cs:6.5,4.5) node[
    scale=0.9,
    text=darkslategray38,
    rotate=0.0
  ]{0};
  \draw (axis cs:7.5,4.5) node[
    scale=0.9,
    text=darkslategray38,
    rotate=0.0
  ]{15};
  \draw (axis cs:8.5,4.5) node[
    scale=0.9,
    text=darkslategray38,
    rotate=0.0
  ]{0};
  \draw (axis cs:0.5,5.5) node[
    scale=0.9,
    text=darkslategray38,
    rotate=0.0
  ]{0};
  \draw (axis cs:1.5,5.5) node[
    scale=0.9,
    text=darkslategray38,
    rotate=0.0
  ]{8};
  \draw (axis cs:2.5,5.5) node[
    scale=0.9,
    text=darkslategray38,
    rotate=0.0
  ]{0};
  \draw (axis cs:3.5,5.5) node[
    scale=0.9,
    text=darkslategray38,
    rotate=0.0
  ]{0};
  \draw (axis cs:4.5,5.5) node[
    scale=0.9,
    text=darkslategray38,
    rotate=0.0
  ]{0};
  \draw (axis cs:5.5,5.5) node[
    scale=0.9,
    text=white,
    rotate=0.0
  ]{392};
  \draw (axis cs:6.5,5.5) node[
    scale=0.9,
    text=darkslategray38,
    rotate=0.0
  ]{0};
  \draw (axis cs:7.5,5.5) node[
    scale=0.9,
    text=darkslategray38,
    rotate=0.0
  ]{0};
  \draw (axis cs:8.5,5.5) node[
    scale=0.9,
    text=darkslategray38,
    rotate=0.0
  ]{0};
  \draw (axis cs:0.5,6.5) node[
    scale=0.9,
    text=darkslategray38,
    rotate=0.0
  ]{0};
  \draw (axis cs:1.5,6.5) node[
    scale=0.9,
    text=darkslategray38,
    rotate=0.0
  ]{0};
  \draw (axis cs:2.5,6.5) node[
    scale=0.9,
    text=darkslategray38,
    rotate=0.0
  ]{0};
  \draw (axis cs:3.5,6.5) node[
    scale=0.9,
    text=darkslategray38,
    rotate=0.0
  ]{0};
  \draw (axis cs:4.5,6.5) node[
    scale=0.9,
    text=darkslategray38,
    rotate=0.0
  ]{0};
  \draw (axis cs:5.5,6.5) node[
    scale=0.9,
    text=darkslategray38,
    rotate=0.0
  ]{0};
  \draw (axis cs:6.5,6.5) node[
    scale=0.9,
    text=white,
    rotate=0.0
  ]{400};
  \draw (axis cs:7.5,6.5) node[
    scale=0.9,
    text=darkslategray38,
    rotate=0.0
  ]{0};
  \draw (axis cs:8.5,6.5) node[
    scale=0.9,
    text=darkslategray38,
    rotate=0.0
  ]{0};
  \draw (axis cs:0.5,7.5) node[
    scale=0.9,
    text=darkslategray38,
    rotate=0.0
  ]{0};
  \draw (axis cs:1.5,7.5) node[
    scale=0.9,
    text=darkslategray38,
    rotate=0.0
  ]{0};
  \draw (axis cs:2.5,7.5) node[
    scale=0.9,
    text=darkslategray38,
    rotate=0.0
  ]{0};
  \draw (axis cs:3.5,7.5) node[
    scale=0.9,
    text=darkslategray38,
    rotate=0.0
  ]{0};
  \draw (axis cs:4.5,7.5) node[
    scale=0.9,
    text=darkslategray38,
    rotate=0.0
  ]{0};
  \draw (axis cs:5.5,7.5) node[
    scale=0.9,
    text=darkslategray38,
    rotate=0.0
  ]{0};
  \draw (axis cs:6.5,7.5) node[
    scale=0.9,
    text=darkslategray38,
    rotate=0.0
  ]{0};
  \draw (axis cs:7.5,7.5) node[
    scale=0.9,
    text=white,
    rotate=0.0
  ]{400};
  \draw (axis cs:8.5,7.5) node[
    scale=0.9,
    text=darkslategray38,
    rotate=0.0
  ]{0};
  \draw (axis cs:0.5,8.5) node[
    scale=0.9,
    text=darkslategray38,
    rotate=0.0
  ]{0};
  \draw (axis cs:1.5,8.5) node[
    scale=0.9,
    text=darkslategray38,
    rotate=0.0
  ]{0};
  \draw (axis cs:2.5,8.5) node[
    scale=0.9,
    text=darkslategray38,
    rotate=0.0
  ]{35};
  \draw (axis cs:3.5,8.5) node[
    scale=0.9,
    text=darkslategray38,
    rotate=0.0
  ]{0};
  \draw (axis cs:4.5,8.5) node[
    scale=0.9,
    text=darkslategray38,
    rotate=0.0
  ]{0};
  \draw (axis cs:5.5,8.5) node[
    scale=0.9,
    text=darkslategray38,
    rotate=0.0
  ]{0};
  \draw (axis cs:6.5,8.5) node[
    scale=0.9,
    text=darkslategray38,
    rotate=0.0
  ]{0};
  \draw (axis cs:7.5,8.5) node[
    scale=0.9,
    text=darkslategray38,
    rotate=0.0
  ]{10};
  \draw (axis cs:8.5,8.5) node[
    scale=0.9,
    text=white,
    rotate=0.0
  ]{355};
  \end{axis}
  
  \end{tikzpicture}
  

%% file: graphs/mlp_confusion_matrix.tex
\begin{tikzpicture}

  \definecolor{darkgray176}{RGB}{176,176,176}
  \definecolor{darkslategray38}{RGB}{38,38,38}
  
  \begin{axis}[
  tick align=outside,
  tick pos=left,
  x grid style={darkgray176},
  xlabel={Predicted Label},
  xmin=0, xmax=9,
  xtick style={color=black},
  xtick={0.5,1.5,2.5,3.5,4.5,5.5,6.5,7.5,8.5},
  xticklabels={0,1,2,3,4,5,6,7,8},
  y dir=reverse,
  y grid style={darkgray176},
  ylabel={True Label},
  ymin=0, ymax=9,
  ytick style={color=black},
  ytick={0.5,1.5,2.5,3.5,4.5,5.5,6.5,7.5,8.5},
  yticklabel style={rotate=90.0},
  yticklabels={0,1,2,3,4,5,6,7,8}
  ]
  \addplot graphics [includegraphics cmd=\pgfimage,xmin=0, xmax=9, ymin=9, ymax=0] {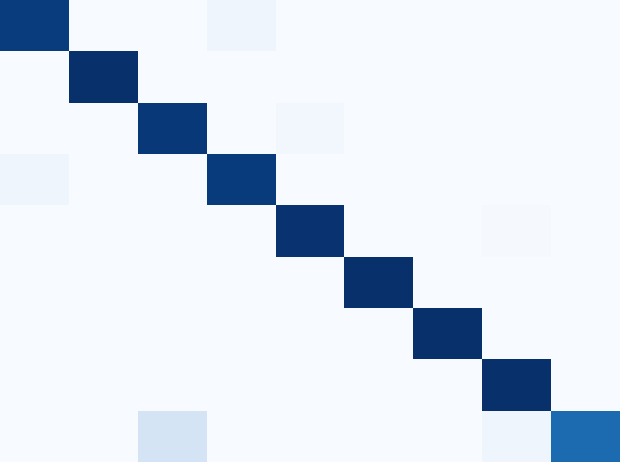};
  \draw (axis cs:0.5,0.5) node[
    scale=0.9,
    text=white,
    rotate=0.0
  ]{381};
  \draw (axis cs:1.5,0.5) node[
    scale=0.9,
    text=darkslategray38,
    rotate=0.0
  ]{0};
  \draw (axis cs:2.5,0.5) node[
    scale=0.9,
    text=darkslategray38,
    rotate=0.0
  ]{0};
  \draw (axis cs:3.5,0.5) node[
    scale=0.9,
    text=darkslategray38,
    rotate=0.0
  ]{19};
  \draw (axis cs:4.5,0.5) node[
    scale=0.9,
    text=darkslategray38,
    rotate=0.0
  ]{0};
  \draw (axis cs:5.5,0.5) node[
    scale=0.9,
    text=darkslategray38,
    rotate=0.0
  ]{0};
  \draw (axis cs:6.5,0.5) node[
    scale=0.9,
    text=darkslategray38,
    rotate=0.0
  ]{0};
  \draw (axis cs:7.5,0.5) node[
    scale=0.9,
    text=darkslategray38,
    rotate=0.0
  ]{0};
  \draw (axis cs:8.5,0.5) node[
    scale=0.9,
    text=darkslategray38,
    rotate=0.0
  ]{0};
  \draw (axis cs:0.5,1.5) node[
    scale=0.9,
    text=darkslategray38,
    rotate=0.0
  ]{0};
  \draw (axis cs:1.5,1.5) node[
    scale=0.9,
    text=white,
    rotate=0.0
  ]{400};
  \draw (axis cs:2.5,1.5) node[
    scale=0.9,
    text=darkslategray38,
    rotate=0.0
  ]{0};
  \draw (axis cs:3.5,1.5) node[
    scale=0.9,
    text=darkslategray38,
    rotate=0.0
  ]{0};
  \draw (axis cs:4.5,1.5) node[
    scale=0.9,
    text=darkslategray38,
    rotate=0.0
  ]{0};
  \draw (axis cs:5.5,1.5) node[
    scale=0.9,
    text=darkslategray38,
    rotate=0.0
  ]{0};
  \draw (axis cs:6.5,1.5) node[
    scale=0.9,
    text=darkslategray38,
    rotate=0.0
  ]{0};
  \draw (axis cs:7.5,1.5) node[
    scale=0.9,
    text=darkslategray38,
    rotate=0.0
  ]{0};
  \draw (axis cs:8.5,1.5) node[
    scale=0.9,
    text=darkslategray38,
    rotate=0.0
  ]{0};
  \draw (axis cs:0.5,2.5) node[
    scale=0.9,
    text=darkslategray38,
    rotate=0.0
  ]{0};
  \draw (axis cs:1.5,2.5) node[
    scale=0.9,
    text=darkslategray38,
    rotate=0.0
  ]{0};
  \draw (axis cs:2.5,2.5) node[
    scale=0.9,
    text=white,
    rotate=0.0
  ]{387};
  \draw (axis cs:3.5,2.5) node[
    scale=0.9,
    text=darkslategray38,
    rotate=0.0
  ]{0};
  \draw (axis cs:4.5,2.5) node[
    scale=0.9,
    text=darkslategray38,
    rotate=0.0
  ]{13};
  \draw (axis cs:5.5,2.5) node[
    scale=0.9,
    text=darkslategray38,
    rotate=0.0
  ]{0};
  \draw (axis cs:6.5,2.5) node[
    scale=0.9,
    text=darkslategray38,
    rotate=0.0
  ]{0};
  \draw (axis cs:7.5,2.5) node[
    scale=0.9,
    text=darkslategray38,
    rotate=0.0
  ]{0};
  \draw (axis cs:8.5,2.5) node[
    scale=0.9,
    text=darkslategray38,
    rotate=0.0
  ]{0};
  \draw (axis cs:0.5,3.5) node[
    scale=0.9,
    text=darkslategray38,
    rotate=0.0
  ]{18};
  \draw (axis cs:1.5,3.5) node[
    scale=0.9,
    text=darkslategray38,
    rotate=0.0
  ]{0};
  \draw (axis cs:2.5,3.5) node[
    scale=0.9,
    text=darkslategray38,
    rotate=0.0
  ]{0};
  \draw (axis cs:3.5,3.5) node[
    scale=0.9,
    text=white,
    rotate=0.0
  ]{382};
  \draw (axis cs:4.5,3.5) node[
    scale=0.9,
    text=darkslategray38,
    rotate=0.0
  ]{0};
  \draw (axis cs:5.5,3.5) node[
    scale=0.9,
    text=darkslategray38,
    rotate=0.0
  ]{0};
  \draw (axis cs:6.5,3.5) node[
    scale=0.9,
    text=darkslategray38,
    rotate=0.0
  ]{0};
  \draw (axis cs:7.5,3.5) node[
    scale=0.9,
    text=darkslategray38,
    rotate=0.0
  ]{0};
  \draw (axis cs:8.5,3.5) node[
    scale=0.9,
    text=darkslategray38,
    rotate=0.0
  ]{0};
  \draw (axis cs:0.5,4.5) node[
    scale=0.9,
    text=darkslategray38,
    rotate=0.0
  ]{0};
  \draw (axis cs:1.5,4.5) node[
    scale=0.9,
    text=darkslategray38,
    rotate=0.0
  ]{0};
  \draw (axis cs:2.5,4.5) node[
    scale=0.9,
    text=darkslategray38,
    rotate=0.0
  ]{0};
  \draw (axis cs:3.5,4.5) node[
    scale=0.9,
    text=darkslategray38,
    rotate=0.0
  ]{0};
  \draw (axis cs:4.5,4.5) node[
    scale=0.9,
    text=white,
    rotate=0.0
  ]{395};
  \draw (axis cs:5.5,4.5) node[
    scale=0.9,
    text=darkslategray38,
    rotate=0.0
  ]{0};
  \draw (axis cs:6.5,4.5) node[
    scale=0.9,
    text=darkslategray38,
    rotate=0.0
  ]{0};
  \draw (axis cs:7.5,4.5) node[
    scale=0.9,
    text=darkslategray38,
    rotate=0.0
  ]{5};
  \draw (axis cs:8.5,4.5) node[
    scale=0.9,
    text=darkslategray38,
    rotate=0.0
  ]{0};
  \draw (axis cs:0.5,5.5) node[
    scale=0.9,
    text=darkslategray38,
    rotate=0.0
  ]{0};
  \draw (axis cs:1.5,5.5) node[
    scale=0.9,
    text=darkslategray38,
    rotate=0.0
  ]{0};
  \draw (axis cs:2.5,5.5) node[
    scale=0.9,
    text=darkslategray38,
    rotate=0.0
  ]{0};
  \draw (axis cs:3.5,5.5) node[
    scale=0.9,
    text=darkslategray38,
    rotate=0.0
  ]{0};
  \draw (axis cs:4.5,5.5) node[
    scale=0.9,
    text=darkslategray38,
    rotate=0.0
  ]{0};
  \draw (axis cs:5.5,5.5) node[
    scale=0.9,
    text=white,
    rotate=0.0
  ]{400};
  \draw (axis cs:6.5,5.5) node[
    scale=0.9,
    text=darkslategray38,
    rotate=0.0
  ]{0};
  \draw (axis cs:7.5,5.5) node[
    scale=0.9,
    text=darkslategray38,
    rotate=0.0
  ]{0};
  \draw (axis cs:8.5,5.5) node[
    scale=0.9,
    text=darkslategray38,
    rotate=0.0
  ]{0};
  \draw (axis cs:0.5,6.5) node[
    scale=0.9,
    text=darkslategray38,
    rotate=0.0
  ]{0};
  \draw (axis cs:1.5,6.5) node[
    scale=0.9,
    text=darkslategray38,
    rotate=0.0
  ]{0};
  \draw (axis cs:2.5,6.5) node[
    scale=0.9,
    text=darkslategray38,
    rotate=0.0
  ]{0};
  \draw (axis cs:3.5,6.5) node[
    scale=0.9,
    text=darkslategray38,
    rotate=0.0
  ]{0};
  \draw (axis cs:4.5,6.5) node[
    scale=0.9,
    text=darkslategray38,
    rotate=0.0
  ]{0};
  \draw (axis cs:5.5,6.5) node[
    scale=0.9,
    text=darkslategray38,
    rotate=0.0
  ]{0};
  \draw (axis cs:6.5,6.5) node[
    scale=0.9,
    text=white,
    rotate=0.0
  ]{400};
  \draw (axis cs:7.5,6.5) node[
    scale=0.9,
    text=darkslategray38,
    rotate=0.0
  ]{0};
  \draw (axis cs:8.5,6.5) node[
    scale=0.9,
    text=darkslategray38,
    rotate=0.0
  ]{0};
  \draw (axis cs:0.5,7.5) node[
    scale=0.9,
    text=darkslategray38,
    rotate=0.0
  ]{0};
  \draw (axis cs:1.5,7.5) node[
    scale=0.9,
    text=darkslategray38,
    rotate=0.0
  ]{0};
  \draw (axis cs:2.5,7.5) node[
    scale=0.9,
    text=darkslategray38,
    rotate=0.0
  ]{0};
  \draw (axis cs:3.5,7.5) node[
    scale=0.9,
    text=darkslategray38,
    rotate=0.0
  ]{0};
  \draw (axis cs:4.5,7.5) node[
    scale=0.9,
    text=darkslategray38,
    rotate=0.0
  ]{0};
  \draw (axis cs:5.5,7.5) node[
    scale=0.9,
    text=darkslategray38,
    rotate=0.0
  ]{0};
  \draw (axis cs:6.5,7.5) node[
    scale=0.9,
    text=darkslategray38,
    rotate=0.0
  ]{0};
  \draw (axis cs:7.5,7.5) node[
    scale=0.9,
    text=white,
    rotate=0.0
  ]{400};
  \draw (axis cs:8.5,7.5) node[
    scale=0.9,
    text=darkslategray38,
    rotate=0.0
  ]{0};
  \draw (axis cs:0.5,8.5) node[
    scale=0.9,
    text=darkslategray38,
    rotate=0.0
  ]{0};
  \draw (axis cs:1.5,8.5) node[
    scale=0.9,
    text=darkslategray38,
    rotate=0.0
  ]{0};
  \draw (axis cs:2.5,8.5) node[
    scale=0.9,
    text=darkslategray38,
    rotate=0.0
  ]{71};
  \draw (axis cs:3.5,8.5) node[
    scale=0.9,
    text=darkslategray38,
    rotate=0.0
  ]{0};
  \draw (axis cs:4.5,8.5) node[
    scale=0.9,
    text=darkslategray38,
    rotate=0.0
  ]{0};
  \draw (axis cs:5.5,8.5) node[
    scale=0.9,
    text=darkslategray38,
    rotate=0.0
  ]{0};
  \draw (axis cs:6.5,8.5) node[
    scale=0.9,
    text=darkslategray38,
    rotate=0.0
  ]{0};
  \draw (axis cs:7.5,8.5) node[
    scale=0.9,
    text=darkslategray38,
    rotate=0.0
  ]{20};
  \draw (axis cs:8.5,8.5) node[
    scale=0.9,
    text=white,
    rotate=0.0
  ]{309};
  \end{axis}
  
  \end{tikzpicture}
  

%% file: graphs/main_table_balanced_accuracy.tex
\begin{tikzpicture}

\definecolor{darkgray176}{RGB}{176,176,176}
\definecolor{darkslategray38}{RGB}{38,38,38}

\begin{groupplot}[group style={group size=4 by 1, horizontal sep=0.4cm, vertical sep=2cm}, width=5.3cm, height=2.8cm]
\nextgroupplot[
axis x line=top,
tick align=outside,
title={5nodes},
title style={yshift=2em},
x grid style={darkgray176},
xmin=0, xmax=3,
xtick pos=right,
xtick style={draw=none}, 
xtick={0.5,1.5,2.5},
xticklabels={KNN,SVM,MLP},
y dir=reverse,
ylabel style={yshift=0em},
ylabel=\parbox{2cm}{\centering Balanced\\Accuracy},
ymin=0, ymax=1,
ytick pos=left,
axis line style={draw=none},
ytick=\empty
]
\addplot graphics [includegraphics cmd=\pgfimage,xmin=0, xmax=3, ymin=1, ymax=0] {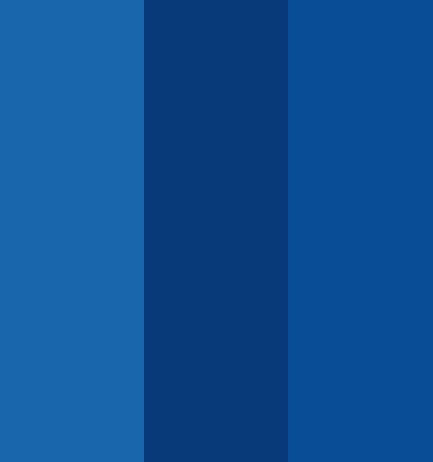};
\draw (axis cs:0.5,0.5) node[
  scale=1.0,
  text=white,
  rotate=0.0
]{0.79};
\draw (axis cs:1.5,0.5) node[
  scale=1.0,
  text=white,
  rotate=0.0
]{0.96};
\draw (axis cs:2.5,0.5) node[
  scale=1.0,
  text=white,
  rotate=0.0
]{0.89};

\nextgroupplot[
axis x line=top,
tick align=outside,
title={4nodes},
title style={yshift=2em},
x grid style={darkgray176},
xmin=0, xmax=3,
xtick pos=right,
xtick style={draw=none}, 
xtick={0.5,1.5,2.5},
xticklabels={KNN,SVM,MLP},
y dir=reverse,
ymin=0, ymax=1,
axis line style={draw=none},  
ytick=\empty
]
\addplot graphics [includegraphics cmd=\pgfimage,xmin=0, xmax=3, ymin=1, ymax=0] {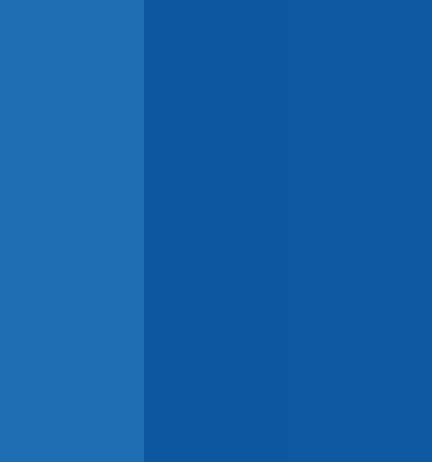};
\draw (axis cs:0.5,0.5) node[
  scale=1.0,
  text=white,
  rotate=0.0
]{0.76};
\draw (axis cs:1.5,0.5) node[
  scale=1.0,
  text=white,
  rotate=0.0
]{0.85};
\draw (axis cs:2.5,0.5) node[
  scale=1.0,
  text=white,
  rotate=0.0
]{0.84};

\nextgroupplot[
axis x line=top,
tick align=outside,
title={3nodes},
title style={yshift=2em},
x grid style={darkgray176},
xmin=0, xmax=3,
xtick pos=right,
xtick style={draw=none}, 
xtick={0.5,1.5,2.5},
xticklabels={KNN,SVM,MLP},
y dir=reverse,
ymin=0, ymax=1,
axis line style={draw=none},  
ytick=\empty
]
\addplot graphics [includegraphics cmd=\pgfimage,xmin=0, xmax=3, ymin=1, ymax=0] {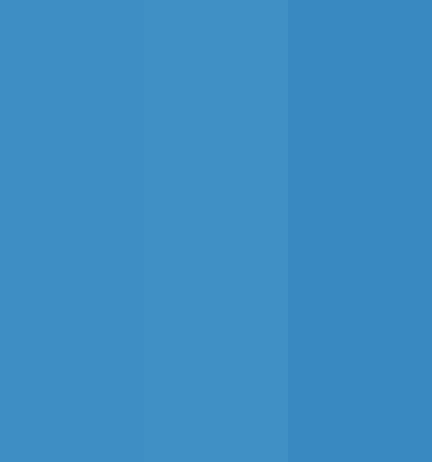};
\draw (axis cs:0.5,0.5) node[
  scale=1.0,
  text=white,
  rotate=0.0
]{0.64};
\draw (axis cs:1.5,0.5) node[
  scale=1.0,
  text=white,
  rotate=0.0
]{0.63};
\draw (axis cs:2.5,0.5) node[
  scale=1.0,
  text=white,
  rotate=0.0
]{0.66};

\nextgroupplot[
axis x line=top,
colorbar,
colorbar style={
    ylabel={},
    draw=none,           
    major tick length=0, 
    minor tick length=0, 
    tick style={draw=none}, 
    ticklabel style={font=\scriptsize}, 
    ytick={0, 1},        
    yticklabels={0, 1},  
    axis line style={draw=none}
},
colormap={mymap}{[1pt]
  rgb(0pt)=(0.968627450980392,0.984313725490196,1);
  rgb(1pt)=(0.870588235294118,0.92156862745098,0.968627450980392);
  rgb(2pt)=(0.776470588235294,0.858823529411765,0.937254901960784);
  rgb(3pt)=(0.619607843137255,0.792156862745098,0.882352941176471);
  rgb(4pt)=(0.419607843137255,0.682352941176471,0.83921568627451);
  rgb(5pt)=(0.258823529411765,0.572549019607843,0.776470588235294);
  rgb(6pt)=(0.129411764705882,0.443137254901961,0.709803921568627);
  rgb(7pt)=(0.0313725490196078,0.317647058823529,0.611764705882353);
  rgb(8pt)=(0.0313725490196078,0.188235294117647,0.419607843137255)
},
point meta max=1,
point meta min=0,
axis x line=top,
tick align=outside,
title={2nodes},
title style={yshift=2em},
x grid style={darkgray176},
xmin=0, xmax=3,
xtick pos=right,
xtick style={draw=none}, 
xtick={0.5,1.5,2.5},
xticklabels={KNN,SVM,MLP},
y dir=reverse,
ymin=0, ymax=1,
axis line style={draw=none},  
ytick=\empty
]
\addplot graphics [includegraphics cmd=\pgfimage,xmin=0, xmax=3, ymin=1, ymax=0] {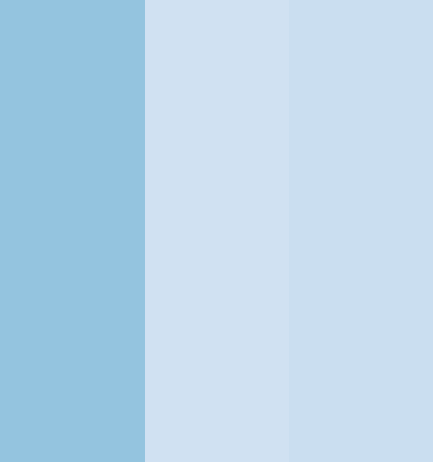};
\draw (axis cs:0.5,0.5) node[
  scale=1.0,
  text=darkslategray38,
  rotate=0.0
]{0.40};
\draw (axis cs:1.5,0.5) node[
  scale=1.0,
  text=darkslategray38,
  rotate=0.0
]{0.20};
\draw (axis cs:2.5,0.5) node[
  scale=1.0,
  text=darkslategray38,
  rotate=0.0
]{0.23};
\end{groupplot}

\end{tikzpicture}

%% file: sec/04_Experiments.tex

\section{Experimental Results}




In this section, we analyze the performance of the KNN model presented in the previous section, as it exhibited the lowest performance among the tested models. This evaluation aims to demonstrate the robustness and practicality of our approach in real-world scenarios. By focusing on the KNN model, we can highlight the challenges and limitations that arise in practical applications. To provide a comprehensive assessment, we conducted experiments in two different real-world scenarios. The first scenario involved using one single drone, while the second scenario included both drone and ground robot. Implementing the KNN model in these varied setups allowed us to observe its effectiveness and identify potential areas for improvement. For our posture prediction, we defined nine distinct classes for the drone: 0 for standby, 1 for up, 2 for down, 3 for takeoff, 4 for land, 5 for left, 6 for right, 7 for forward, and 8 for backward movement. For the ground robot, we used only classes 0, 5, 6, 7, and 8.

\subsection{Experimental Setup}

The experimental platform for this study comprises a commercially available Ryze Tello MAV and Turtlebot4 platform both equipped with OptiTrack motion capture (MOCAP) system markers. This setup enables precise recording of the robot's trajectory while executing movement commands, allowing us to monitor their movements and verify the accuracy of our model predictions.

For ground truth, we utilized a MOCAP system consisting of six OptiTrack cameras connected to a backbone WiFi network. This system covers an operating area approximately 8 meters wide, 9 meters long, and 5 meters high.

In this experiment, a separate individual, distinct from those who provided training and testing data for the model, was equipped with five DWM1001-Dev UWB nodes programmed with custom firmware for measuring the distances~\cite{moron2022towards}. These nodes were strategically placed on the belly, wrists, and ankles to ensure comprehensive data collection. 

The primary system for training the models and making predictions operates on Ubuntu 20.04. All nodes responsible for receiving UWB data, estimating postures, and sending movement commands to the robots were running ROS~2 Galactic. All the recorded data from our experiments are available in the project's repository\footnote{https://github.com/salmasalimii/UWB-based-posture-recognition}.

\subsection{Obtained Results}

To demonstrate the versatility and capability of mapping a wide array of human postures or classes to either specific degrees of freedom in robots or to more actions in complex robots and multi-robot systems, we conducted two real-world experiments. The first experiment involved controlling only a Tello drone, while the second involved a Tello drone and a TurtleBot4 high-level teleoperation. In both experiments, a KNN model was used for human pose estimation.

\begin{figure}[tb]
    \centering
    \includegraphics[width=0.5\textwidth]{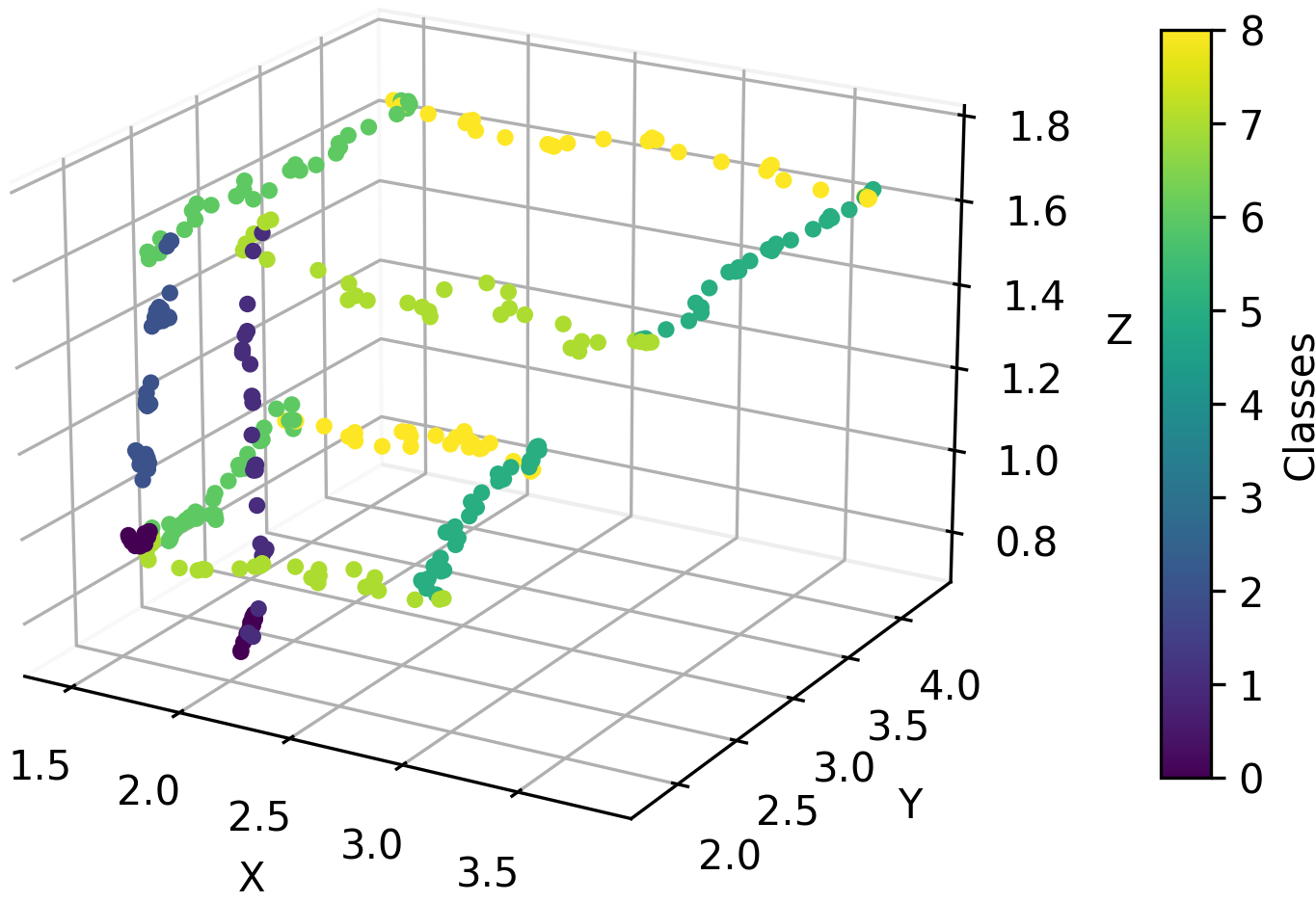}  
    \caption{Drone 3D path trajectory}
    \label{fig:3d_tello_path}
\end{figure}

As shown in Fig.~\ref{fig:3d_tello_path}, during the experiment utilizing the KNN model for posture estimation, the participant successfully maneuvered the drone within two designated squares at varying heights. The image illustrates the 3D movement of the drone, highlighting its ability to navigate through space accurately. Additionally, the 3D plot includes a color bar indicating the posture predicted at each movement, demonstrating that the predictions were quite accurate and allowed the participant to control the drone with ease. This successful demonstration underscores the robustness and practical applicability of our system in real-world scenarios. The ability to accurately interpret human postures and translate them into precise drone movements underscores the potential of our approach for various applications in human-robot interaction and autonomous systems.

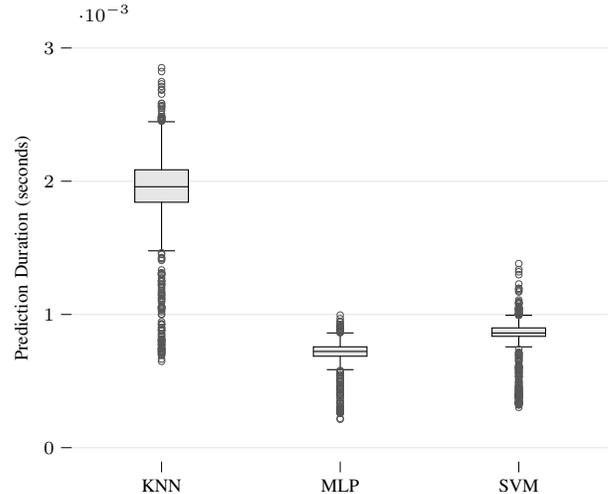
\begin{figure}[tb]
    \centering
    \setlength\figureheight{0.4\textwidth}
    \setlength\figurewidth{0.48\textwidth}
    \scriptsize{\input{graphs/prediction_duration_boxplot}}
    \caption{Model prediction duration throughout the experiment}
    \label{fig:prediction_duration_plot}
\end{figure}

Furthermore, in Fig.~\ref{fig:prediction_duration_plot}, we calculated the prediction duration during our experiment to assess the latency in prediction calculation across all three models. For the KNN model, the prediction duration was around 0.002 seconds on average, while for SVM and MLP, it was less than 0.001 seconds. This analysis highlights the efficiency of our models, particularly the KNN model, in providing timely and accurate posture predictions, which is crucial for real-time control in autonomous systems.

\begin{figure}[tb]
    \centering
    \setlength\figureheight{0.29\textwidth}
    \setlength\figurewidth{0.48\textwidth}
    \scriptsize{\input{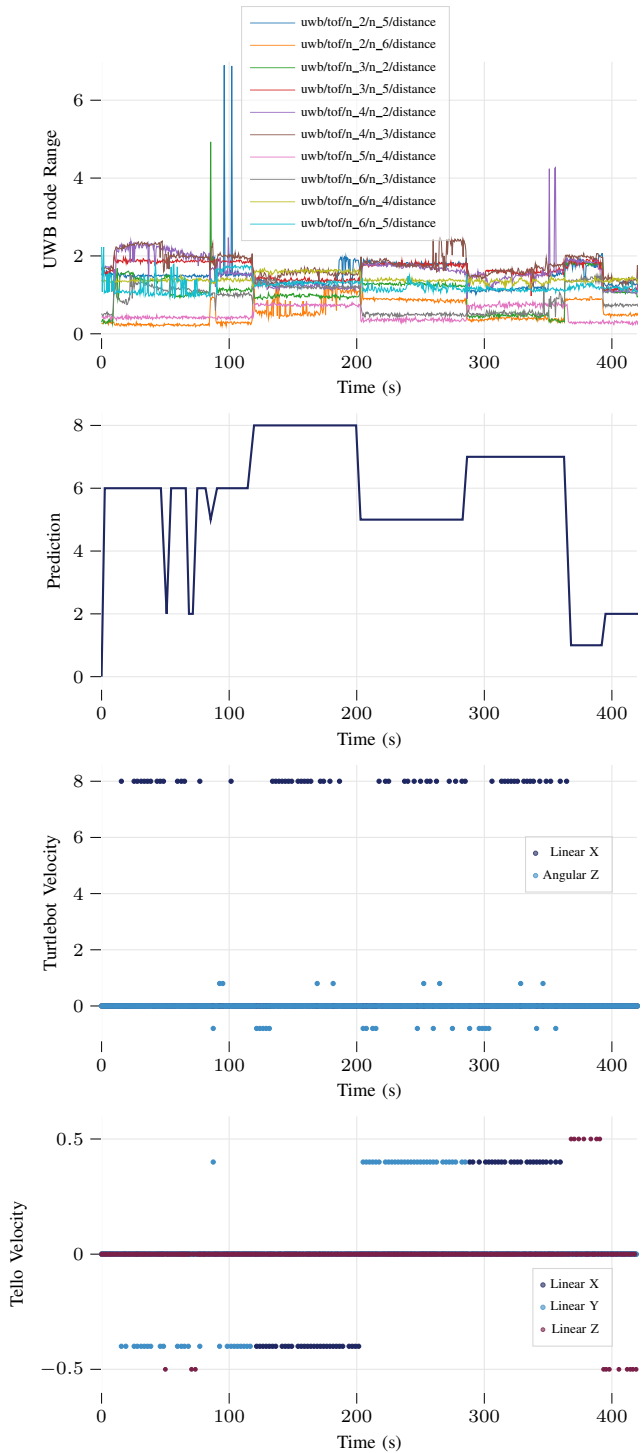}}
    \caption{Second scenario experiment results}
    \label{fig:combined_plot}
\end{figure}

In the second scenario, we employed two robots: one Tello drone and one TurtleBot. Fig.~\ref{fig:combined_plot} illustrates the UWB node ranges during the experiment, along with the posture predictions made by the KNN model. For clearer visualization, the linear and angular velocity commands sent to the robots are also depicted. During this experiment, the participant maneuvered the robots in a square pattern and adjusted the drone's altitude. Given that the TurtleBot's height is fixed, certain postures were used exclusively to command the drone's vertical movements.

\begin{figure}[tb]
    \centering
    \setlength\figureheight{0.4\textwidth}
    \setlength\figurewidth{0.5\textwidth}
    \scriptsize{\input{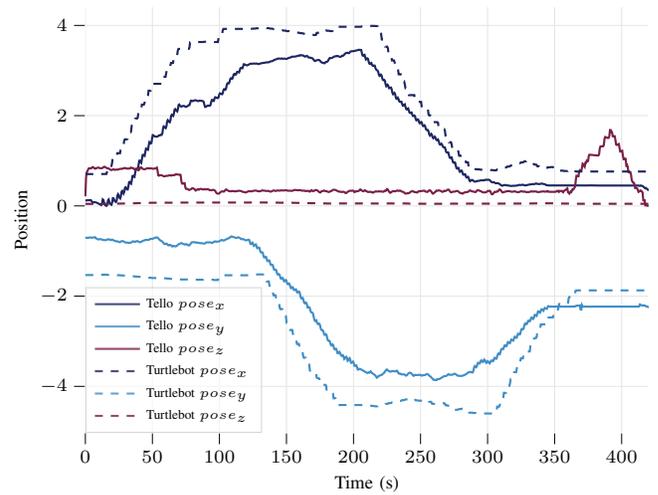}}
    \caption{Tello and Turtlebot trajectory}
    \label{fig:tello_pose_plot}
\end{figure}

Despite encountering significant noise at certain points, as shown in the UWB ranges plot, the KNN model maintained sufficient accuracy to effectively control the robots. This demonstrates the robustness of our approach in handling real-world noise and ensuring reliable performance. Fig.~\ref{fig:tello_pose_plot} depicts the trajectory tracking of both robots, indicating that our model accurately predicted the postures, allowing the participant to guide the robots as intended. The successful coordination between the drone and the TurtleBot highlights the model's capability to facilitate precise and responsive control in multi-robot systems, underscoring its potential for practical applications in complex environments.




%% file: graphs/prediction_duration_boxplot.tex
\begin{tikzpicture}

\definecolor{darkgray}{RGB}{86, 86, 86}
\definecolor{darkorange25512714}{RGB}{255,127,14}
\definecolor{steelblue31119180}{RGB}{31,119,180}

\definecolor{color0}{rgb}{0.9,0.9,0.9}
\definecolor{color1}{HTML}{1E2761}
\definecolor{color2}{HTML}{408EC6}
\definecolor{color3}{HTML}{7A2048}

\begin{axis}[
height=\figureheight,
width=\figurewidth,
axis line style={white},
legend style={fill opacity=0.8, draw opacity=1, text opacity=1, 
draw=white!80!black},
tick align=outside,
tick pos=left,
x grid style={white!69.0196078431373!black},
xmin=0.5, xmax=3.5,
xtick style={color=black},
xtick={1,2,3},
xticklabels={KNN,MLP,SVM},
y grid style={white!90!black},
ymajorgrids,
ylabel={Prediction Duration (seconds)},
ymin=-0.0001, ymax=0.0031,
ytick style={color=black}
]
\path [draw=black, fill=color0]
(axis cs:0.85,0.0018422603607177)
--(axis cs:1.15,0.0018422603607177)
--(axis cs:1.15,0.0020859241485595)
--(axis cs:0.85,0.0020859241485595)
--(axis cs:0.85,0.0018422603607177)
--cycle;
\addplot [black]
table {%
1 0.0018422603607177
1 0.0014777183532714
};
\addplot [black]
table {%
1 0.0020859241485595
1 0.0024466514587402
};
\addplot [black]
table {%
0.925 0.0014777183532714
1.075 0.0014777183532714
};
\addplot [black]
table {%
0.925 0.0024466514587402
1.075 0.0024466514587402
};
\addplot [darkgray, mark=o, mark size=1.23, mark options={solid,fill opacity=0}, only marks]
table {%
1 0.0008835792541503
1 0.0013110637664794
1 0.0008127689361572
1 0.0010914802551269
1 0.0008022785186767
1 0.0007517337799072
1 0.0008127689361572
1 0.0008783340454101
1 0.0008063316345214
1 0.0009455680847167
1 0.0007991790771484
1 0.0007963180541992
1 0.0012543201446533
1 0.0007798671722412
1 0.0008018016815185
1 0.0011441707611083
1 0.0008494853973388
1 0.0009222030639648
1 0.0009255409240722
1 0.0008172988891601
1 0.000694990158081
1 0.0011599063873291
1 0.0007777214050292
1 0.0007438659667968
1 0.001002550125122
1 0.0013039112091064
1 0.0007233619689941
1 0.0007274150848388
1 0.0011830329895019
1 0.001103401184082
1 0.0012252330780029
1 0.000786542892456
1 0.0007414817810058
1 0.0012476444244384
1 0.0007300376892089
1 0.0013341903686523
1 0.0014052391052246
1 0.0014636516571044
1 0.0013008117675781
1 0.0009305477142333
1 0.000716209411621
1 0.0014166831970214
1 0.0014219284057617
1 0.0008561611175537
1 0.000899314880371
1 0.0011126995086669
1 0.0009982585906982
1 0.0011253356933593
1 0.0012555122375488
1 0.0011529922485351
1 0.0010440349578857
1 0.0009009838104248
1 0.001049518585205
1 0.001033067703247
1 0.0008866786956787
1 0.0011441707611083
1 0.0012276172637939
1 0.0013017654418945
1 0.0010552406311035
1 0.0011210441589355
1 0.0010497570037841
1 0.0006656646728515
1 0.0007150173187255
1 0.0010623931884765
1 0.0012261867523193
1 0.0007061958312988
1 0.0007693767547607
1 0.0014503002166748
1 0.0012149810791015
1 0.0010581016540527
1 0.0008308887481689
1 0.0008761882781982
1 0.0012454986572265
1 0.0013134479522705
1 0.0007381439208984
1 0.0012745857238769
1 0.0011680126190185
1 0.0007705688476562
1 0.0011999607086181
1 0.0007462501525878
1 0.0010211467742919
1 0.0012943744659423
1 0.0007646083831787
1 0.0008389949798583
1 0.0010430812835693
1 0.0007426738739013
1 0.0007102489471435
1 0.0006482601165771
1 0.0006959438323974
1 0.000725507736206
1 0.001427412033081
1 0.002568244934082
1 0.0025844573974609
1 0.0024545192718505
1 0.0028231143951416
1 0.0024526119232177
1 0.0024831295013427
1 0.0025830268859863
1 0.0025486946105957
1 0.0027475357055664
1 0.0025320053100585
1 0.0024666786193847
1 0.002730369567871
1 0.0027114650726318
1 0.0026838779449462
1 0.0024855136871337
1 0.0025031566619873
1 0.0024831295013427
1 0.0024800300598144
1 0.0024535655975341
1 0.0024707317352294
1 0.0025603771209716
1 0.0028519630432128
1 0.0024659633636474
1 0.0025639533996582
1 0.0024538040161132
1 0.0024752616882324
1 0.0026540756225585
1 0.0026884078979492
1 0.0024757385253906
1 0.00246262550354
};
\path [draw=black, fill=color0]
(axis cs:1.85,0.0006866455078125)
--(axis cs:2.15,0.0006866455078125)
--(axis cs:2.15,0.0007572174072265)
--(axis cs:1.85,0.0007572174072265)
--(axis cs:1.85,0.0006866455078125)
--cycle;
\addplot [black]
table {%
2 0.0006866455078125
2 0.0005857944488525
};
\addplot [black]
table {%
2 0.0007572174072265
2 0.0008614063262939
};
\addplot [black]
table {%
1.925 0.0005857944488525
2.075 0.0005857944488525
};
\addplot [black]
table {%
1.925 0.0008614063262939
2.075 0.0008614063262939
};
\addplot [darkgray, mark=o, mark size=1.23, mark options={solid,fill opacity=0}, only marks]
table {%
2 0.0002593994140625
2 0.0002615451812744
2 0.0002877712249755
2 0.0002133846282958
2 0.0002996921539306
2 0.0003318786621093
2 0.0004034042358398
2 0.0002839565277099
2 0.0004355907440185
2 0.0005297660827636
2 0.0002779960632324
2 0.0004422664642333
2 0.0003952980041503
2 0.0002198219299316
2 0.000279426574707
2 0.0003039836883544
2 0.0005137920379638
2 0.0004994869232177
2 0.0005671977996826
2 0.0004591941833496
2 0.00044846534729
2 0.0004994869232177
2 0.0004653930664062
2 0.0004971027374267
2 0.0005781650543212
2 0.0004968643188476
2 0.0005462169647216
2 0.0003483295440673
2 0.0004951953887939
2 0.0005159378051757
2 0.0005779266357421
2 0.0002868175506591
2 0.0005044937133789
2 0.0004637241363525
2 0.0005419254302978
2 0.0002884864807128
2 0.0002684593200683
2 0.0004053115844726
2 0.0003376007080078
2 0.0003437995910644
2 0.0003764629364013
2 0.0002591609954833
2 0.0003693103790283
2 0.0002937316894531
2 0.0002827644348144
2 0.0004401206970214
2 0.0005412101745605
2 0.0004010200500488
2 0.0003042221069335
2 0.0003976821899414
2 0.0005676746368408
2 0.0002739429473876
2 0.0002877712249755
2 0.000270128250122
2 0.0002665519714355
2 0.0004289150238037
2 0.0002756118774414
2 0.0002758502960205
2 0.000274658203125
2 0.000277042388916
2 0.0002715587615966
2 0.0005633831024169
2 0.0003144741058349
2 0.0004777908325195
2 0.0002937316894531
2 0.0004181861877441
2 0.0002741813659667
2 0.0009081363677978
2 0.0009477138519287
2 0.000925064086914
2 0.000995397567749
2 0.0009117126464843
2 0.0008773803710937
2 0.0009181499481201
2 0.0008695125579833
2 0.000864028930664
2 0.0009045600891113
2 0.0008735656738281
2 0.0008692741394042
2 0.0008723735809326
2 0.0008697509765625
2 0.0008876323699951
2 0.0008893013000488
2 0.0008835792541503
2 0.0008974075317382
2 0.0008704662322998
2 0.0008764266967773
2 0.0008707046508789
2 0.0009682178497314
2 0.0008645057678222
2 0.0008726119995117
2 0.0009412765502929
2 0.000910997390747
};
\path [draw=black, fill=color0]
(axis cs:2.85,0.00083577632904045)
--(axis cs:3.15,0.00083577632904045)
--(axis cs:3.15,0.000899314880371)
--(axis cs:2.85,0.000899314880371)
--(axis cs:2.85,0.00083577632904045)
--cycle;
\addplot [black]
table {%
3 0.00083577632904045
3 0.0007569789886474
};
\addplot [black]
table {%
3 0.000899314880371
3 0.0009942054748535
};
\addplot [black]
table {%
2.925 0.0007569789886474
3.075 0.0007569789886474
};
\addplot [black]
table {%
2.925 0.0009942054748535
3.075 0.0009942054748535
};
\addplot [darkgray, mark=o, mark size=1.23, mark options={solid,fill opacity=0}, only marks]
table {%
3 0.0006434917449951
3 0.0003449916839599
3 0.000303030014038
3 0.0003392696380615
3 0.0003206729888916
3 0.0005545616149902
3 0.0005445480346679
3 0.0004551410675048
3 0.0003359317779541
3 0.0003523826599121
3 0.0005743503570556
3 0.0007145404815673
3 0.0003385543823242
3 0.0004422664642333
3 0.0004236698150634
3 0.0006065368652343
3 0.0004441738128662
3 0.0003314018249511
3 0.000333547592163
3 0.0003430843353271
3 0.0003395080566406
3 0.000619888305664
3 0.000596284866333
3 0.0004329681396484
3 0.0006895065307617
3 0.0004100799560546
3 0.0007154941558837
3 0.0006630420684814
3 0.0003299713134765
3 0.0003881454467773
3 0.0003323554992675
3 0.0003833770751953
3 0.0007083415985107
3 0.0006020069122314
3 0.0003845691680908
3 0.0004565715789794
3 0.0004880428314208
3 0.0003745555877685
3 0.0003752708435058
3 0.0004322528839111
3 0.0004022121429443
3 0.000380516052246
3 0.0003988742828369
3 0.0003383159637451
3 0.0004940032958984
3 0.0006077289581298
3 0.0005085468292236
3 0.0003895759582519
3 0.0005970001220703
3 0.0004236698150634
3 0.0006937980651855
3 0.0003972053527832
3 0.0003974437713623
3 0.0004112720489501
3 0.0003819465637207
3 0.0005161762237548
3 0.0005357265472412
3 0.0003440380096435
3 0.0007071495056152
3 0.0003936290740966
3 0.0003774166107177
3 0.0003314018249511
3 0.0004403591156005
3 0.0006089210510253
3 0.0003886222839355
3 0.0003798007965087
3 0.0004663467407226
3 0.0003824234008789
3 0.0003595352172851
3 0.0003750324249267
3 0.0003960132598876
3 0.0006101131439208
3 0.0004994869232177
3 0.0003621578216552
3 0.0004060268402099
3 0.00040864944458
3 0.0005273818969726
3 0.0006752014160156
3 0.0004880428314208
3 0.0003912448883056
3 0.0005936622619628
3 0.0004317760467529
3 0.0003762245178222
3 0.0004665851593017
3 0.0003802776336669
3 0.0003941059112548
3 0.0003783702850341
3 0.0004210472106933
3 0.0004630088806152
3 0.0004737377166748
3 0.0003824234008789
3 0.0003790855407714
3 0.000373363494873
3 0.0003471374511718
3 0.0003759860992431
3 0.0006783008575439
3 0.0004253387451171
3 0.0004053115844726
3 0.0003817081451416
3 0.0003409385681152
3 0.0004007816314697
3 0.0004050731658935
3 0.00042724609375
3 0.0006499290466308
3 0.0005946159362792
3 0.0003881454467773
3 0.0004010200500488
3 0.0003666877746582
3 0.0004827976226806
3 0.0003764629364013
3 0.0005357265472412
3 0.0004041194915771
3 0.0004043579101562
3 0.0006635189056396
3 0.0007016658782958
3 0.000582218170166
3 0.0004215240478515
3 0.0003693103790283
3 0.0003812313079833
3 0.0004537105560302
3 0.0003836154937744
3 0.0005939006805419
3 0.0004410743713378
3 0.0004336833953857
3 0.0003249645233154
3 0.0005559921264648
3 0.0005784034729003
3 0.000328779220581
3 0.0003368854522705
3 0.0003364086151123
3 0.000328779220581
3 0.0005309581756591
3 0.0005617141723632
3 0.0003345012664794
3 0.0003371238708496
3 0.0003471374511718
3 0.000338077545166
3 0.0005645751953125
3 0.0003271102905273
3 0.0004501342773437
3 0.0006918907165527
3 0.0005805492401123
3 0.0003235340118408
3 0.0003321170806884
3 0.0006768703460693
3 0.0003345012664794
3 0.0006229877471923
3 0.0004537105560302
3 0.0006530284881591
3 0.0005366802215576
3 0.0005655288696289
3 0.0003337860107421
3 0.0007383823394775
3 0.0010817050933837
3 0.0010013580322265
3 0.0010969638824462
3 0.0012989044189453
3 0.0010108947753906
3 0.0010433197021484
3 0.0010032653808593
3 0.0010037422180175
3 0.0011940002441406
3 0.0010437965393066
3 0.0010056495666503
3 0.0010309219360351
3 0.0013377666473388
3 0.0010240077972412
3 0.0009956359863281
3 0.0010898113250732
3 0.0011970996856689
3 0.0011875629425048
3 0.0009958744049072
3 0.0010032653808593
3 0.0009975433349609
3 0.0009949207305908
3 0.001169204711914
3 0.0010292530059814
3 0.0010037422180175
3 0.0010526180267333
3 0.0010476112365722
3 0.0013816356658935
3 0.0010409355163574
3 0.0010087490081787
3 0.0011088848114013
3 0.0011775493621826
3 0.0013203620910644
3 0.0010032653808593
3 0.0010240077972412
3 0.0010154247283935
3 0.00108003616333
3 0.001082420349121
3 0.0010435581207275
3 0.0010371208190917
3 0.0012290477752685
3 0.0010633468627929
3 0.0010879039764404
3 0.0010368824005126
3 0.0010433197021484
3 0.0009994506835937
3 0.0009949207305908
3 0.0010502338409423
};
\addplot [black]
table {%
0.85 0.0019586086273193
1.15 0.0019586086273193
};
\addplot [black]
table {%
1.85 0.0007224082946777
2.15 0.0007224082946777
};
\addplot [black]
table {%
2.85 0.00086104869842525
3.15 0.00086104869842525
};
\end{axis}

\end{tikzpicture}

%% file: sec/05_Discussion.tex

\section{Discussion}

In this study, we have conducted a comprehensive comparison of the KNN, SVM, and MLP models, particularly evaluating their latency and performance in posture prediction tasks. The analysis highlights the strengths and weaknesses of each model in different contexts such as performance under noisy conditions, computational latency, and scalability for real-time applications. Furthermore, two real-world experiments were carried out using the KNN model to further assess its effectiveness. The results of these experiments demonstrate the practical applicability of our approach, offering valuable insights into its performance under real-world conditions. This analysis provides a basis for understanding the trade-offs and identifying potential areas for further refinement.


%% file: sec/06_Conclusion.tex

\section{Conclusion}\label{sec:conclusion}


We have presented an evaluation of three machine learning models—KNN, SVM, and MLP—for human posture estimation. Our study included a detailed real-world examination of KNN's performance, which generally showed slightly lower accuracy compared to SVM and MLP. Across real-world experiments involving two scenarios with multi-robots, KNN's performance was analyzed, highlighting its competitive accuracy and efficiency in predicting human postures. In both controlled settings and real-world applications, our approach demonstrated practical utility and responsiveness, showing variations in performance across different posture classes and in the presence of noise. Despite these challenges, our approach's adaptability to noise and environmental factors underscores its robustness, making it suitable for applications requiring precise human-robot interaction and autonomous system control.